%% file: main.tex
\documentclass[sigconf]{acmart}

\usepackage{multirow}

\usepackage{array}
\usepackage{booktabs}
\usepackage{pifont}
\usepackage{xcolor}

\AtBeginDocument{%
  }

\copyrightyear{2026}
\acmYear{2026}
\setcopyright{cc}
\setcctype{by}
\acmConference[KDD '26]{Proceedings of the 32nd ACM SIGKDD Conference on Knowledge Discovery and Data Mining V.2}{August 09--13, 2026}{Jeju Island, Republic of Korea}
\acmBooktitle{Proceedings of the 32nd ACM SIGKDD Conference on Knowledge Discovery and Data Mining V.2 (KDD '26), August 09--13, 2026, Jeju Island, Republic of Korea}
\acmDOI{10.1145/3770855.3817505}
\acmISBN{979-8-4007-2259-2/2026/08}

\begin{document}

\title{SurveyReview: A Reviewer-Aligned Benchmark for Survey Evaluators}


\author{Yuheng Zhang}
\authornote{These authors contributed equally.} 
\authornote{Work was done when these authors interned at Zhipu AI}
\orcid{0009-0005-5664-968X}
\affiliation{%
  \department{School of Information}
  \institution{Renmin University of China}
  \city{Beijing}
  \country{China}
}
\affiliation{
  \institution{Key Laboratory of Data Engineering and Knowledge Engineering}
  \city{Beijing}
  \country{China}
}
\email{zhangyuheng@ruc.edu.cn}

\author{Yuanchun Wang}
\authornotemark[1]
\authornotemark[2]
\affiliation{%
  \department{School of Information}
  \institution{Renmin University of China}
  \city{Beijing}
  \country{China}
}
\affiliation{
  \institution{Key Laboratory of Data Engineering and Knowledge Engineering}
  \city{Beijing}
  \country{China}
}
\email{wangyuanchun@ruc.edu.cn}

\author{Fanjin Zhang}
\authornotemark[1]
\orcid{0000-0001-8551-1966}
\affiliation{%
  \department{School of Information}
  \institution{Renmin University of China}
  \city{Beijing}
  \country{China}
}
\affiliation{%
  \institution{Engineering Research Center of Database and Business Intelligence}
  \city{Beijing}
  \country{China}
}
\email{fanjinz@ruc.edu.cn}

\author{Ruyu Zhao}
\authornotemark[2]
\orcid{0009-0003-9426-4911}
\affiliation{%
  \department{School of Information}
  \institution{Renmin University of China}
  \city{Beijing}
  \country{China}
}
\affiliation{
  \institution{Key Laboratory of Data Engineering and Knowledge Engineering}
  \city{Beijing}
  \country{China}
}
\email{ruyuzhao@ruc.edu.cn}

\author{Juanzi Li}
\orcid{0000-0002-6244-0664}
\author{Jie Tang}
\orcid{0000-0003-3487-4593}
\affiliation{%
  \department{Department of Computer Science and Technology}
  \institution{Tsinghua University}
  \city{Beijing}
  \country{China}
}
\email{lijuanzi@tsinghua.edu.cn}
\email{jietang@tsinghua.edu.cn}

\author{Jing Zhang}
\authornote{Corresponding author.}
\orcid{0000-0003-2019-225X}
\affiliation{%
  \department{School of Information}
  \institution{Renmin University of China}
  \city{Beijing}
  \country{China}
}
\affiliation{%
  \institution{Engineering Research Center of Database and Business Intelligence}
  \city{Beijing}
  \country{China}
}
\email{zhang-jing@ruc.edu.cn}


\renewcommand{\shortauthors}{Yuheng Zhang et al.}

\begin{abstract}
The rapid advancement of large language models has transformed survey writing from a months-long manual effort into an automated process. As generation scales, reliable evaluation becomes the bottleneck, and LLMs are increasingly used as survey evaluators. However, existing approaches largely rely on off-the-shelf LLM-as-a-judge methods without systematic alignment to human reviewers, and there remains a lack of systematic frameworks for quantifying alignment with human reviewers.
To address this gap, we propose SurveyReview, a reviewer-aligned, multi-dimensional benchmark and dataset for survey evaluation. We collect and annotate 675 survey papers with 1,630 review reports. We structure authentic peer-review reports by converting free-form comments into four-dimensional scores (Readability, Criticalness, Comprehensiveness, Structure) paired with supporting rationales.  We further release standardized train/test splits and an evaluation protocol to measure alignment between automatic evaluators and human reviewers.
To validate the benchmark, we develop SurveyAlign, a strong baseline evaluator by fine-tuning Qwen3-32B with LoRA on our annotated data, 
augmented with external knowledge for knowledge-intensive dimensions. On the test set, SurveyAlign 
substantially improves reviewer alignment over prompt-based judging with GPT-5.2, reducing average MSE from 2.28 to 1.38 and MAE from 1.15 to 0.69 across all four dimensions.
Our contributions are twofold: (1) we establish the first multi-dimensional, reviewer-aligned dataset with a reproducible evaluation framework for survey reviewing; (2) we develop a strong baseline evaluator that substantially improves alignment with human reviewers, providing a competitive reference for future research. 
Our code and data are available at  \url{https://surveyreview.github.io}
\end{abstract}


  

\maketitle


\begin{figure}[t]
    \centering
    \includegraphics[width=\linewidth]{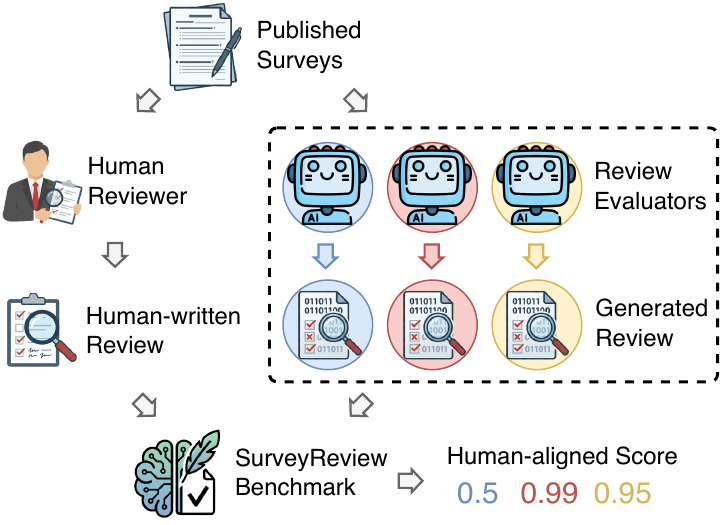}
    \caption{SurveyReview Benchmark: Evaluating to what extent the reviews produced by existing survey evaluators align with those of human expert reviewers.}
    \label{fig:intro1}
    \Description{A diagram illustrating the SurveyReview benchmark}
\end{figure}

\section{Introduction}

\input{Sections/Introduction}

\section{Related Work}

\input{Sections/RelatedWorks}

\section{Task Formulation}
\label{sec:taskformulation}
\input{Sections/Taskformulation}


\section{Dataset Construction}

\input{Sections/Datasetconstruction}

\section{Methods}
\label{sec:benchmark_methods}
\input{Sections/Benchmarkmethods}

\section{Experiments and Analysis}

\input{Sections/Experiments}

\section{Conclusion}

\input{Sections/Conclusion}


\section{Ethical Considerations}
\input{Sections/Ethical}


\begin{acks}
This work is supported by the National Key Research \& Development Plan (2023YFF0725100), the National Natural Science Foundation of China (92570121, 62322214, U23A20299, U24B20144, 62406164, 62476150, 62425601), a grant from the Institute for Guo Qiang, Tsinghua University (2019GQB0003),
and the new cornerstone Science Foundation through the XPLORER PRIZE.
We also acknowledge the support from Public Computing Cloud, Renmin University of China.
We acknowledge the data annotators at Zhipu AI for their great support in this work. 

\end{acks}

\bibliographystyle{ACM-Reference-Format}
\bibliography{bibliography}


\appendix

\input{Sections/Appendix}

\end{document}

%% file: Sections/Introduction.tex
Large language models (LLMs) have transformed survey paper generation from a months-long manual process into an automated pipeline that retrieves literature~\cite{zhang2024oag}, structures content~\cite{zhu2025context}, and drafts comprehensive manuscripts within hours~\cite{wang2024autosurvey,chen2025surveygen,liang2025surveyx}. 
Systems such as DeepResearch~\cite{du2025deepresearchbenchcomprehensivebenchmark,li2025reportbenchevaluatingdeepresearch} orchestrate retrieval, clustering, outline generation, and multi-round refinement to produce surveys that appear fluent and comprehensive in coverage. 
However, this automation creates a critical evaluation bottleneck: while generation scales rapidly, quality assessment remains a major bottleneck and still often relies on human peer review~\cite{chen2025surveygen}. 
Although LLMs are increasingly used as automatic evaluators~\cite{2024automatic}, existing approaches largely apply off-the-shelf \emph{LLM-as-a-judge} prompting~\cite{zheng2023judgingllmasajudgemtbenchchatbot} without systematic alignment to expert reviewers, and there is no rigorous framework to quantify such alignment.


An effective framework must satisfy three requirements that human reviewers naturally provide: (1) multi-dimensional assessment across stable quality dimensions (e.g., readability, structure, comprehensiveness, criticalness); (2) interpretable rationales that explain why each score was assigned, enabling targeted revision; and (3) quantifiable alignment with human judgments, measured via error metrics (MSE, MAE) to reveal systematic biases and ensure reliability. Simply producing scores is insufficient---without structured rationales and measurable alignment, automated evaluators cannot support iterative improvement or establish credibility.

Prior work evaluates surveys from reference-based perspectives rather than directly modeling expert peer review. 
SurveyScope~\cite{shi2025scisage} formalizes evaluation through reference-based comparison across curated survey datasets. 
SurGE~\cite{su2026surge} and DeepSurvey-Bench~\cite{zhang2026deepsurveybench} compare generated surveys against expert-written references, scoring dimensions like coverage, citation accuracy, and ``academic value'' via LLM-as-a-judge. 
While valuable, these approaches treat expert surveys as gold-standard texts rather than learning from expert review reports. 
Consequently, they lack: (1) quality dimensions grounded in actual reviewing practice across diverse topics; (2) structured ``score + rationale'' representations that unify expert judgments into learnable signals; and (3) explicit alignment quantification via MSE/MAE to measure deviation from human reviewers. 
No prior work constructs evaluation datasets from peer review reports or systematically measures alignment error across quality dimensions.

We take a reviewer-aligned perspective and build an evaluation framework directly from real peer-review data. Specifically, we propose \textbf{SurveyReview}, a reviewer-aligned, multi-dimensional benchmark and dataset for survey evaluation. We collect and annotate 675 survey papers with 1,630 review reports, deriving four core dimensions consistently emphasized by experts: readability, structure, comprehensiveness, and criticalness. Readability captures clarity and ease of understanding;
Structure measures the logical organization and coherence of sections;
Comprehensiveness evaluates the breadth and relevance of literature coverage;
Criticalness assesses analytical depth and the extent of original insight beyond summarization. We convert free-form review comments into structured dimension-specific scores paired with supporting rationales, 
and release standardized splits and an evaluation protocol for reproducible alignment measurement. 
Figure~\ref{fig:intro1} illustrates the overall SurveyReview benchmark, 
where reviews generated by survey evaluators are compared with human-written reviews 
to compute human-aligned scores.

Building upon this dataset, we develop \textbf{SurveyAlign}, 
an evaluator based on Qwen3-32B with LoRA adaptation. 
In addition to standard supervised fine-tuning, 
we incorporate reference-expansion-based Knowledge Augmentation for knowledge-intensive dimensions~\cite{zhang2019oag}.
This design reflects the role of references in expert review 
and strengthens coverage and analytical assessment.

We evaluate automatic evaluators along two aspects: score alignment and reasoning quality. Alignment with human reviewers is measured using MSE and MAE across dimensions, while rationale quality is assessed based on semantic consistency with reference explanations. We further integrate these components into a unified \textbf{Human-aligned Score (HAS)}, 
which jointly reflects scoring accuracy and reasoning fidelity with respect to expert peer review.
Experiments show that SurveyAlign substantially improves reviewer alignment, reducing average MSE from 2.28 to 1.38 and MAE from 1.15 to 0.69 compared with GPT-5.2.

We summarize our key contributions as follows:

\paragraph{Dataset}
We establish the first multi-dimensional, reviewer-aligned dataset derived from peer review reports. Under a unified rubric, each paper receives fine-grained scores and structured rationales across four dimensions (readability, structure, comprehensiveness, criticalness). We also provide standardized train/test splits and an evaluation protocol to measure reviewer alignment in a reproducible way.

\paragraph{Evaluator}
We develop SurveyAlign, a strong baseline evaluator built on Qwen3-32B with LoRA.
Beyond text-only fine-tuning, we incorporate a modular knowledge augmentation toolkit that enables structured integration of relevant domain knowledge for knowledge-intensive dimensions such as comprehensiveness and criticalness.
On the test set, SurveyAlign reduces average MSE from 2.28 to 1.38 and MAE from 1.15 to 0.69 compared with GPT-5.2.

%% file: Sections/RelatedWorks.tex
\subsection{Survey Generation}

Automatic survey generation has attracted increasing attention as the volume of scientific literature continues to grow. 
Existing systems primarily focus on improving long-form generation pipelines through retrieval, planning, and multi-stage writing.

AutoSurvey~\cite{wang2024autosurvey} presents an end-to-end pipeline combining literature retrieval, outline construction, and section-wise generation, highlighting evaluation as a critical bottleneck. 
SurveyGen~\cite{chen2025surveygen} introduces a planning-based framework with memory-guided writing to improve global consistency in long survey articles. 
SurveyX~\cite{liang2025surveyx} addresses context limitations via structured knowledge organization and multi-stage generation to enhance coverage and citation accuracy. 
LLM$\times$MapReduce-V2~\cite{wang2025llmmapreduce} further explores test-time scaling strategies for synthesizing long-form academic content from extremely long resources.

While these approaches significantly advance generation quality, 
evaluation is often not the primary focus, 
often relying on coverage heuristics or LLM-based judgment.

\subsection{Survey Evaluation}

Existing work on survey evaluation can be categorized according to the primary source of evaluation signals: 
(1) LLM-centered scoring, 
(2) reference-grounded comparison, and 
(3) task-driven assessment.

\textbf{LLM-centered scoring} relies on large language models as primary evaluators to assess generated surveys according to predefined rubrics. 
SurGE~\cite{su2026surge} and DeepSurvey-Bench~\cite{zhang2026deepsurveybench} adopt rubric-guided LLM scoring, 
sometimes complemented with auxiliary automatic metrics or reference-based statistics. 
While scalable, these approaches predominantly depend on model-generated judgments rather than structured supervision derived from authentic peer-review annotations.

\textbf{Reference-grounded evaluation} measures quality relative to expert-written surveys. 
SurveyScope (SciSage)~\cite{shi2025scisage} evaluates generated surveys against curated references in terms of content coverage and citation usage. 
Traditional metrics such as ROUGE~\cite{lin-2004-rouge}, BLEU~\cite{2022bleu}, and BERTScore~\cite{zhang2020bertscoreevaluatingtextgeneration}, are also applied to long-form academic text~\cite{2024automatic}.

\textbf{Task-driven assessment} evaluates research agents through structured academic tasks rather than direct survey comparison. 
DeepResearch Bench~\cite{du2025deepresearchbenchcomprehensivebenchmark} and ReportBench~\cite{li2025reportbenchevaluatingdeepresearch} assess systems via expert-designed report-level protocols and task completion criteria.

Despite their scalability, these paradigms primarily derive supervision from model judgments, reference texts, or task outcome signals, rather than from real-world peer-review annotations.
They do not explicitly model authentic expert peer-review behavior, nor do they provide dimension-level reviewer scores paired with rationales for alignment-based measurement. 
In contrast, SurveyReview derives structured evaluation signals from authentic peer-review reports and quantifies reviewer-aligned error across dimension-level scores paired with real reviewer rationales.

%% file: Sections/Taskformulation.tex
We formalize survey evaluation as a multi-dimensional scoring task 
that mimics expert peer review. Given a survey paper $\mathcal{P}$, 
our goal is to produce a structured evaluation $\mathcal{E}$ aligning 
with expert judgments:

\begin{equation}
\mathcal{E} = f(\mathcal{P}) = \{(d_i, s_i, r_i)\}_{i=1}^{n}
\end{equation}

where $d_i \in \mathcal{D}$ is a quality dimension, $s_i$ is the 
numeric score, $r_i$ is the supporting rationale, and $n$ is the 
number of dimensions.

\paragraph{Evaluation protocol.}
We measure evaluator alignment with human reviewers using annotated 
surveys as ground truth. For each survey paper $\mathcal{P}$, the corresponding expert evaluation is denoted as 
$\mathcal{E}^* = \{(d_i, s_i^*, r_i^*)\}_{i=1}^{n}$, 
where $s_i^*$ and $r_i^*$ represent the ground-truth score and rationale for dimension $d_i$. 

For dimension $d_i$ with $m_i$ samples, we compute mean squared error and mean absolute error between human scores 
$\{s_{i,j}^*\}_{j=1}^{m_i}$ 
and predictions 
$\{\hat{s}_{i,j}\}_{j=1}^{m_i}$.

\begin{equation}
\text{MSE}_i = \frac{1}{m_i}\sum_{j=1}^{m_i}(s_{i,j}^* - \hat{s}_{i,j})^2, \quad
\text{MAE}_i = \frac{1}{m_i}\sum_{j=1}^{m_i}|s_{i,j}^* - \hat{s}_{i,j}|
\end{equation}

Overall metrics average across all $n$ dimensions:

\begin{equation}
\text{AVGMSE} = \frac{1}{n}\sum_{i=1}^{n}\text{MSE}_i, \quad
\text{AVGMAE} = \frac{1}{n}\sum_{i=1}^{n}\text{MAE}_i
\end{equation}

Additionally, we evaluate reasoning quality by measuring semantic alignment between generated rationales $\{\hat{r}_{i,j}\}$ and reference rationales $\{r_{i,j}^*\}$ via LLM-based assessment. For each dimension $d_i$, we compute the average reason quality score:

\begin{equation}
\text{RQ}_i = \frac{1}{m_i}\sum_{j=1}^{m_i}\text{Consistency}(\hat{r}_{i,j}, r_{i,j}^*)
\end{equation}

where $\text{Consistency}(\cdot, \cdot) \in [0,1]$ measures rationale consistency. The overall reason quality score averages across all dimensions:

\begin{equation}
\text{RQS} = \frac{1}{n}\sum_{i=1}^{n}\text{RQ}_i
\end{equation}

\paragraph{Human-aligned Score (HAS)}
We define the \textbf{Human-aligned Score (HAS)} to combine score alignment and reasoning quality. 
Let $\Delta_s$ denote the maximum possible absolute score difference under our predefined score scale. 
We normalize MSE and MAE into alignment scores:

\begin{equation}
A_{\text{MSE}} = 1 - \frac{\text{AVGMSE}}{\Delta_s^2}, \quad
A_{\text{MAE}} = 1 - \frac{\text{AVGMAE}}{\Delta_s}.
\end{equation}

Higher values indicate better score alignment.

Then compute the weighted alignment and final HAS:

\begin{equation}
\text{Align} = \alpha_1 \cdot A_{\text{MSE}} + \alpha_2 \cdot A_{\text{MAE}}
\end{equation}
\begin{equation}
\text{HAS} = \beta_1 \cdot \text{Align} + \beta_2 \cdot \text{RQS}
\end{equation}


\paragraph{SurveyReview Evaluation.}
In our evaluation protocol, each evaluator is required to output a 
structured four-dimensional assessment for a given survey paper, 
consisting of \textbf{(i)} a numeric score and \textbf{(ii)} a supporting 
rationale for each quality dimension: \textbf{Readability}, 
\textbf{Criticalness}, \textbf{Comprehensiveness}, and \textbf{Structure}.

\textbf{Readability} reflects clarity and ease of understanding;
\textbf{Criticalness} captures analytical depth and synthesized insight beyond summarization;
\textbf{Comprehensiveness} evaluates the breadth and relevance of literature coverage;
and \textbf{Structure} measures the logical organization and coherence of the survey.

Score alignment is measured by AVGMSE and AVGMAE, while rationale 
quality is quantified by the Reason Quality Score (RQS). 
The final Human-aligned Score (HAS) integrates both components 
into a unified metric that quantifies overall alignment with expert peer review.

%% file: Sections/Datasetconstruction.tex
\begin{figure*}[ht]
  \centering
  \includegraphics[width=\linewidth]{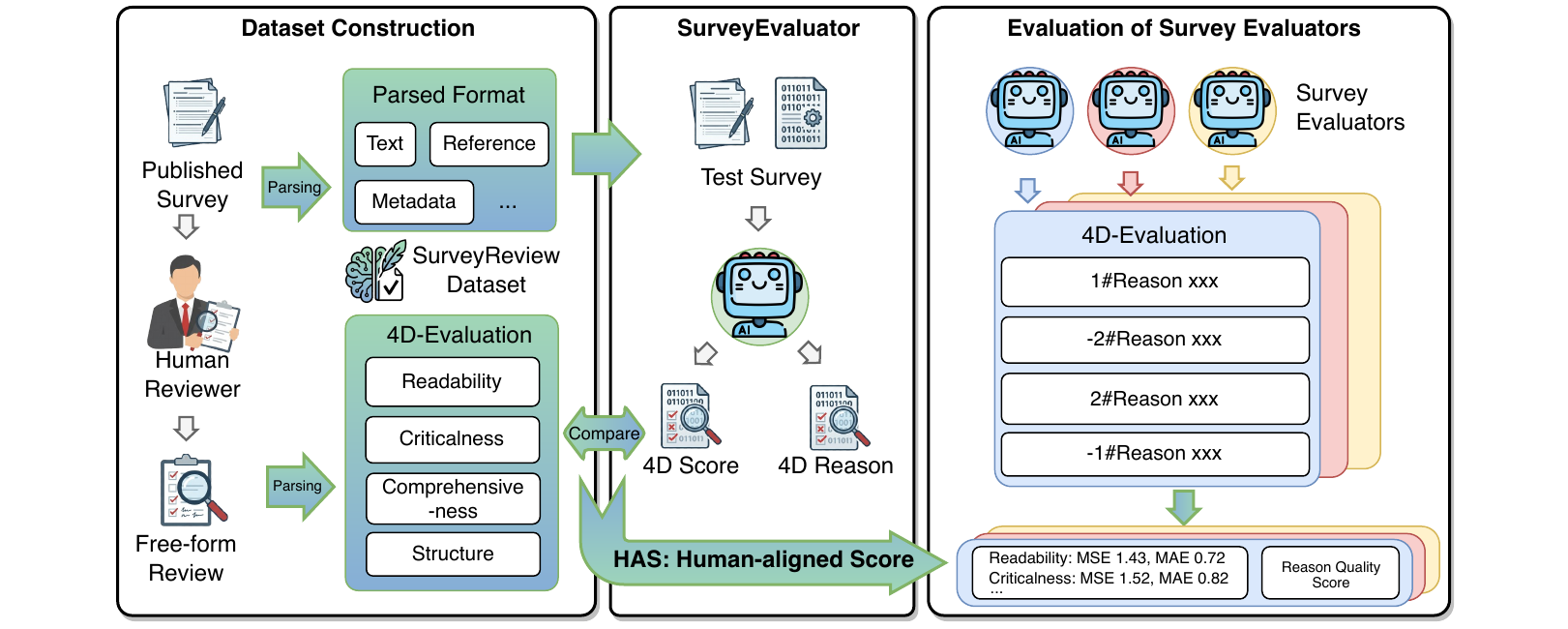}
\caption{Overview of SurveyReview. 
Left: dataset construction. 
Middle: the workflow of a single survey evaluator producing dimension-specific score–rationale pairs. 
Right: evaluation of survey evaluators under the unified benchmarking protocol.}

  \Description{}
  \label{fig:framework}
\end{figure*}

Evaluating automatic survey evaluators requires ground-truth data consisting of paired survey papers and expert assessments, denoted as $(\mathcal{P}, \mathcal{E}^*)$, where $\mathcal{P}$ is a survey paper and $\mathcal{E}^*$ represents the ground-truth four-dimensional scores and supporting rationales provided by human reviewers. Such data serves two key purposes: supervised training and alignment-based benchmarking.

However, collecting high-quality annotations from domain experts at scale is prohibitively expensive~\cite{feng2025sampleefficienthumanevaluationlarge}. While existing peer review processes provide expert evaluations of survey papers, 
these resources are not directly suitable for alignment-based benchmarking. 
To our knowledge, no existing dataset systematically aggregates survey–review pairs across domains with standardized, dimension-level labels. 
Moreover, raw review texts are typically unstructured free-form comments, 
lacking explicit scoring criteria and rationale annotations.

To address these challenges, we construct \textsc{SurveyReview} through a three-stage pipeline. We \textbf{collect} survey papers with peer reviews from multiple academic sources (Section~\ref{subsec:data_collection}), \textbf{process} each PDF to extract full text and citation structure (Section~\ref{subsec:processing}), and \textbf{annotate} reviews into structured four-dimensional scores and rationales (Section~\ref{subsec:annotation}). The resulting dataset contains 675 surveys with multi-dimensional expert annotations, enabling both supervised training and rigorous evaluation. The overall dataset construction pipeline is illustrated on the left side of Figure~\ref{fig:framework}.

\subsection{Data Collection}
\label{subsec:data_collection}

\paragraph{Data sources.}
We collect survey papers and their accompanying peer review reports from three publicly available platforms: MOPRD~\cite{lin2023moprd}, a large-scale multidisciplinary dataset providing structured open peer review records across the full review lifecycle; F1000Research, an open publishing platform adopting a publish-first, post-publication peer review model with fully transparent reviewer reports and evaluation statuses~\cite{thelwall2019doesuseopennonanonymous}; and OpenReview, a widely used transparent peer review and conference management system in the machine learning and AI community. Together, these sources offer diverse domains, review processes, and transparency levels, enabling comprehensive and heterogeneous evaluation settings ~\cite{sun2025openreviewprotectedleveragedcommunity}.

\paragraph{Survey identification.}
For each source, we adopt a three-stage filtering strategy to identify survey papers, consisting of keyword-based matching on titles (e.g., ``survey'' and ``review'') to obtain high-recall candidates ~\cite{zhuang2024understandingsurveypapertaxonomy}, large language model–based screening of titles and abstracts to verify survey-style content~\cite{huotala2025aisysrevllmbasedtool}, and manual verification to remove ambiguous or misclassified cases, thereby ensuring high precision and data quality. We retain only papers that pass all stages and include substantive peer review reports with explicit quality assessments or critical feedback.

\begin{table*}[t]
    \small
    \centering
    \renewcommand\arraystretch{1.0}
    \begin{tabular}{lcccccc}
        \toprule
        Datasets & Fields & \#Surveys & Metadata & Real Reviews & Quality Diversity & Fine-grained Annotation \\
        \midrule
        SurveyScope~\cite{shi2025scisage} \ & CS & 46 
        & \textcolor{green}{\ding{51}} 
        & \textcolor{red}{\ding{55}}  
        & \textcolor{red}{\ding{55}} 
        & \textcolor{red}{\ding{55}} \\

        SurveyBench~\cite{sun2025surveybench} & CS & 100 
        & \textcolor{green}{\ding{51}} 
        & \textcolor{red}{\ding{55}} 
        & \textcolor{red}{\ding{55}} 
        & \textcolor{red}{\ding{55}} \\

        SurGE~\cite{su2026surge} & CS & 205 
        & \textcolor{green}{\ding{51}} 
        & \textcolor{red}{\ding{55}} 
        & \textcolor{red}{\ding{55}} 
        & \textcolor{red}{\ding{55}} \\

        DeepSurvey-Bench~\cite{zhang2026deepsurveybench} & Mixed & 163 
        & \textcolor{green}{\ding{51}} 
        & \textcolor{red}{\ding{55}} 
        & \textcolor{red}{\ding{55}} 
        & \textcolor{red}{\ding{55}} \\
        
        \midrule
        \textbf{SurveyReview} & Mixed & 675
        & \textcolor{green}{\ding{51}} 
        & \textcolor{green}{\ding{51}} 
        & \textcolor{green}{\ding{51}} 
        & \textcolor{green}{\ding{51}} \\
        \bottomrule
    \end{tabular}
    \caption{Comparison of survey datasets. 
    Real Reviews denotes the use of authentic peer review reports. 
    Quality Diversity denotes the availability of both positive and negative quality signals. 
    Fine-grained Annotation denotes structured dimension-level scores with rationales.}
    \label{tab:survey_dataset_compare}
\end{table*}

\subsection{Survey Paper Processing}
\label{subsec:processing}

After identifying survey papers with accompanying reviews, we retrieve the corresponding PDF files for downstream processing and analysis.

\paragraph{PDF parsing.}
We extract structured information from the PDFs using two complementary tools. Marker~\cite{marker} converts PDFs into clean markdown text, providing full paper content for evaluator input. Grobid~\cite{grobid} parses PDFs into structured XML, extracting reference lists, citation contexts, and bibliographic metadata. This dual-tool approach yields comprehensive paper representations, including full text, citation graphs, and metadata.

\paragraph{Extracted information.}
The processing pipeline produces:
\begin{itemize}
    \item Full-text content (via Marker): complete paper text for evaluation
    \item References (via Grobid): reference lists, citation contexts, and locations
    \item Metadata: title, authors, abstract, and publication year
\end{itemize}

This structured representation supports both evaluator training (using full text) and knowledge augmentation (using citation analysis).

\subsection{Annotation Methodology}
\label{subsec:annotation}

We convert free-form peer review reports into structured 
dimension-level annotations. For each survey–review pair, 
annotators assign scores and extract supporting rationales 
for four predefined dimensions: 
\textit{Readability}, \textit{Structure}, 
\textit{Comprehensiveness}, and \textit{Criticalness}. 

The resulting ground-truth expert evaluation for each survey paper 
is denoted as 
$\mathcal{E}^* = \{(d_i, s_i^*, r_i^*)\}_{i=1}^{n}$, 
where $s_i^*$ and $r_i^*$ represent the annotated score and 
corresponding rationale for dimension $d_i$.

Annotation guidelines specify task definition, 
dimension criteria, rationale extraction rules, 
and scoring standards. More annotation details are provided in Appendix~\ref{appendix:annotationdetails}.

All samples are annotated by trained annotators. 
To assess annotation reliability, we randomly sample 280 annotation units 
from the dataset, where each unit corresponds to one evaluation dimension 
of one survey paper. 
This subset is independently annotated by multiple annotators. 
Inter-annotator agreement on this subset, measured by 
Krippendorff's alpha, is 0.74. We provide per-dimension agreement statistics and disagreement pattern 
analysis in Appendix~\ref{appendix:annotation_reliability}.

\subsection{Dataset Statistics}
The SurveyReview dataset comprises 675 survey papers 
and 1,630 associated peer review reports with structured 
four-dimensional annotations.

As summarized in Table~\ref{tab:survey_dataset_compare}, 
SurveyReview differs from existing survey benchmarks in 
three key aspects. First, it is constructed from authentic 
peer review reports rather than synthetic judgments or 
reference-only comparisons. Second, the dataset contains 
both positive and negative quality signals, reflecting 
the naturally critical distribution of real-world reviews. 
Third, each sample is annotated along four predefined 
quality dimensions with explicit supporting rationales, 
enabling fine-grained supervision and evaluation.

\paragraph{Source composition.}
Reviews are collected from F1000Research 
(1,058; 64.9\%), MOPRD (375; 23.0\%), and OpenReview 
(197; 12.1\%), providing heterogeneous domains and 
review styles. 
\paragraph{Score distribution.}
Dimension-wise statistics are reported in Table~\ref{tab:dimension_stats}. 
The overall score distribution is: 
$-2$ (4.46\%), $-1$ (66.25\%), $+1$ (24.73\%), and $+2$ (4.56\%). 
The distribution is skewed toward negative assessments, with extreme scores ($\pm2$) accounting for a small proportion of annotations.
\paragraph{Train–test split.}
The dataset is split into 1,216 training samples (75\%) 
and 414 test samples (25\%), stratified by data source 
to preserve domain proportions. The split is performed 
at the paper level: all reviews associated with the same 
survey paper are assigned to the same partition, ensuring 
that no paper appears in both training and test sets.

%% file: Sections/Benchmarkmethods.tex
Under the task formulation in Section~\ref{sec:taskformulation}, 
survey evaluation aims to learn a mapping 
$f_\theta: \mathcal{P} \rightarrow \hat{\mathcal{E}}$, 
where 
\[
\hat{\mathcal{E}} = f_\theta(\mathcal{P}) 
= \{(d_i, \hat{s}_i, \hat{r}_i)\}_{i=1}^{n}
\]
denotes the predicted dimension-specific score–rationale pairs 
for a survey paper $\mathcal{P}$.

To realize this mapping in practice, we develop 
\textbf{SurveyAlign}, a modular evaluator that implements 
$f_\theta$ under the unified evaluation protocol. 
SurveyAlign combines supervised fine-tuning (SFT) 
for surface-level dimensions 
(\textit{Readability} and \textit{Structure}) 
with dimension-specific Knowledge Augmentation (KA) 
for knowledge-intensive dimensions 
(\textit{Comprehensiveness} and \textit{Criticalness}), 
and further applies inference-time consensus aggregation 
to improve prediction stability.

\subsection{Supervised Fine-Tuning (SFT)}
\label{subsec:sft}

In supervised fine-tuning, we train an instruction-following LLM on 
SurveyReview annotations to learn reviewer-aligned scoring and 
rationale generation. We construct instruction-tuning examples 
consisting of \texttt{instruction}, \texttt{input}, and \texttt{output} fields, 
each corresponding to a single evaluation dimension. 
The \texttt{input} contains the survey text, and the 
\texttt{output} includes a numeric score paired with a 
human-written rationale. We adopt the same dimension-specific prompting scheme used for zero-shot evaluation in Appendix~\ref{appendix:promptllm}, ensuring consistency between prompt-based baselines and supervised fine-tuning.
 
We build dimension-specific training sets using annotated reviews. 
Qwen3-32B~\cite{yang2025qwen3technicalreport}  is fine-tuned with LoRA~\cite{hu2021loralowrankadaptationlarge} for parameter-efficient adaptation, 
learning the mapping from survey text to dimension-specific 
score–rationale pairs.

Implementation details are provided in Appendix~\ref{appendix:sft}.

\subsection{Knowledge-Augmented Training}
\label{subsec:ka_training}

The four evaluation dimensions exhibit different dependency structures. 
\textit{Readability} and \textit{Structure} primarily rely on 
text-internal signals such as fluency and organization. 
In contrast, \textit{Comprehensiveness} and \textit{Criticalness} 
require reasoning beyond the survey text itself. 
Comprehensiveness involves assessing literature coverage relative 
to the surrounding research landscape, while Criticalness requires 
analytical synthesis across multiple works. 
These dimensions therefore benefit from explicit Knowledge Augmentation (KA).

\subsubsection{Reference Expansion}

Given a survey paper $\mathcal{P}$, we extract its reference set $R$. 
Using a citation graph constructed from AMiner~\cite{tang2016aminer}, 
we expand $R$ to obtain an augmented citation neighborhood $R^{+}$ 
by retrieving papers that have citation relationships with items in $R$. 
This expansion provides a structured view of the local research 
landscape surrounding the survey.

\subsubsection{Dimension-Specific Knowledge Augmentation}

Since $R^{+}$ can exceed the available context window, 
we construct compact, dimension-specific KA representations 
before injecting external knowledge into the model input.

\paragraph{Comprehensiveness-specific Knowledge Augmentation (Comp-KA)}
Comprehensiveness focuses on coverage completeness. 
We therefore identify structurally important works in $R^{+}$ 
based on their connectivity to the original reference set $R$. 
Candidate papers are ranked by citation links with $R$, 
and the top-ranked papers are selected together with the original references. 
The resulting Comp-KA representation is serialized as structured 
metadata (title, year, citation count), highlighting representative 
and influential works in the citation neighborhood. 
This allows the model to reason about potential coverage gaps.

\paragraph{Criticalness-specific Knowledge Augmentation (Crit-KA)}
Criticalness requires assessing analytical depth and synthesized insight. 
Unlike coverage evaluation, this dimension depends on modeling 
the evolution and interaction of ideas over time. 
We therefore construct a temporally structured representation from $R^{+}$. 
Papers are filtered using adaptive thresholds derived from 
reference statistics (e.g., citation count and publication year), 
forming a representative candidate set. 
Candidates are sorted chronologically and grouped under a token budget, 
each block summarized, and the summaries further compressed into 
a fixed-length representation (Crit-KA). 
This preserves coarse temporal evolution signals to support 
cross-paper analytical reasoning.

Detailed construction procedures are provided in 
Appendix~\ref{appendix:ka_details}.

\subsubsection{Training with Knowledge-Augmented Inputs}

For knowledge-intensive dimensions, the corresponding KA 
representation is prepended to the survey text as part of the model input.

For \textit{Comprehensiveness}, supervised fine-tuning (SFT) 
is performed on inputs of the form [Comp-KA; $\mathcal{P}$].

For \textit{Criticalness}, we adopt a two-stage strategy. 
First, SFT is conducted on inputs augmented with Crit-KA. 
We then apply Direct Preference Optimization (DPO)~\cite{rafailov2024directpreferenceoptimizationlanguage} 
under the same augmented setting. 
Criticalness judgments are inherently more subjective and require 
nuanced synthesis beyond factual grounding. 
DPO further aligns the model toward human-preferred analytical reasoning.

Separate LoRA-adapted models~\cite{hu2021loralowrankadaptationlarge} 
are trained for each dimension, enabling dimension-specific 
supervision and optimization strategies.

\subsection{Model Composition and Inference}
\label{SuperReviewer}

\paragraph{Test-Time Scaling via Consensus Aggregation.}
To reduce stochastic variance in generative scoring, 
we apply a lightweight test-time aggregation strategy 
inspired by prior work on scaling test-time compute for LLMs~\cite{agarwal2025artscalingtesttimecompute,weng2024cycleresearcher}. 
For each evaluation dimension, the corresponding model is executed 
multiple times with sampling, producing independent score–rationale pairs. 
The final score is determined via majority voting across runs, 
and the rationales are consolidated into a unified explanation 
consistent with the consensus score. 
This strategy introduces no additional training cost 
and improves prediction stability purely through inference-time aggregation. 
Detailed configurations are provided in Appendix~\ref{appendix:testtimescaling}.

\paragraph{Overall Construction.}

SurveyAlign is designed as a modular evaluation framework composed of four 
dimension-specialized sub-models: 
\textit{SurveyAlign-Read}, 
\textit{SurveyAlign-Crit}, 
\textit{SurveyAlign-Comp}, and 
\textit{SurveyAlign-Stru}. 
Each sub-model is independently fine-tuned with LoRA to focus on a single evaluation dimension, 
allowing dimension-specific supervision signals and optimization strategies.

For the surface-level dimensions (\textit{Readability} and \textit{Structure}), 
we adopt supervised fine-tuning (SFT) 
(Section~\ref{subsec:sft}) to improve score calibration and rationale consistency. 
For the knowledge-intensive dimensions 
(\textit{Criticalness} and \textit{Comprehensiveness}), 
we further incorporate knowledge-augmented training 
(Section~\ref{subsec:ka_training}) to enhance evidence grounding and coverage assessment.

At inference time, each dimension-specialized sub-model applies 
the consensus-based test-time scaling strategy described above, 
and the resulting dimension-level score–rationale pairs 
are combined into a unified multi-dimensional evaluation output.

%% file: Sections/Experiments.tex
\begin{table*}[th]
    \caption{Main results across all evaluated methods. AVG: average across the four evaluation dimensions. RQS: Reason Quality Score. HAS: Human-aligned Score}
    \label{tab:main_results}
    \resizebox{\textwidth}{!}{%
    \begin{tabular}{ll*{10}{c}cc}
        \toprule
        & & \multicolumn{2}{c}{\textbf{Read.}} & \multicolumn{2}{c}{\textbf{Crit.}} & \multicolumn{2}{c}{\textbf{Comp.}} & \multicolumn{2}{c}{\textbf{Stru.}} & \multicolumn{2}{c}{\textbf{AVG}} & \multirow{2}{*}{\textbf{RQS} $\uparrow$} & \multirow{2}{*}{\textbf{HAS} $\uparrow$}  \\
        \cmidrule(lr){3-4} \cmidrule(lr){5-6} \cmidrule(lr){7-8} \cmidrule(lr){9-10} \cmidrule(lr){11-12}
        \textbf{Method} & \textbf{Model} & MSE $\downarrow$ & MAE $\downarrow$ & MSE $\downarrow$ & MAE $\downarrow$ & MSE $\downarrow$ & MAE $\downarrow$ & MSE $\downarrow$ & MAE $\downarrow$ & MSE $\downarrow$ & MAE $\downarrow$ & & \\
        \midrule
        SurveyAlign & Qwen3-32B-LoRA & \textbf{1.43} & \textbf{0.72} & \textbf{1.52} & \textbf{0.82} & \textbf{1.26} & \textbf{0.56} & \textbf{1.29} & \textbf{0.65} & \textbf{1.38} & \textbf{0.69} & 0.36 & \textbf{0.74} \\
        \midrule
        \multicolumn{14}{l}{\textit{Zero-shot}} \\
        Prompt & Qwen3-32B & 3.05 & 1.45 & 3.24 & 1.51 & 3.22 & 1.54 & 3.35 & 1.53 & 3.21 & 1.51 & 0.36 & 0.61 \\
        Prompt & GPT-5.2 & 2.13 & 1.07 & 1.97 & 0.97 & 2.04 & 1.08 & 2.98 & 1.47 & 2.28 & 1.15 & 0.42 & 0.68 \\
        Prompt & Gemini-3-pro & 3.84 & 1.52 & 2.25 & 1.00 & 3.91 & 1.49 & 5.76 & 2.11 & 3.94 & 1.53 & 0.29 & 0.58 \\
        Prompt & DeepSeek-v3.2 & 4.78 & 1.88 & 2.49 & 1.15 & 4.59 & 1.82 & 4.02 & 1.76 & 3.97 & 1.65 & 0.37 & 0.58 \\
        Prompt & GLM-4.7 & 3.43 & 1.50 & 2.58 & 1.21 & 3.66 & 1.57 & 4.83 & 1.95 & 3.62 & 1.56 & 0.37 & 0.60 \\
        Prompt & Claude-Opus-4.5 & 2.91 & 1.29 & 1.88 & 0.88 & 2.66 & 1.23 & 3.65 & 1.58 & 2.77 & 1.25 & \textbf{0.48} & 0.68 \\
        \midrule
        \multicolumn{14}{l}{\textit{Prior Work}} \\
        LLMMapReduce & Gemini-3-pro & 2.95 & 1.27 & 3.04 & 1.29 & 5.04 & 1.83 & 6.13 & 2.24 & 4.29 & 1.66 & -- & -- \\
        DR-Bench & GPT-5.2 & 3.06 & 1.39 & 2.51 & 1.13 & 2.75 & 1.29 & -- & -- & -- & -- & -- & -- \\
        SurveyX & GPT-5.2 & -- & -- & 2.14 & 1.00 & -- & -- & 3.19 & 1.43 & -- & -- & -- & -- \\
        \bottomrule
    \end{tabular}%
    }
\end{table*}


\subsection{Experimental Setup}
\label{subsec:exp_setup}

\paragraph{Evaluation Metrics.}
Following the task formulation in Section~\ref{sec:taskformulation}, 
we evaluate automatic survey evaluators along two aspects: 
\emph{score alignment} and \emph{rationale quality}. 

For each evaluation dimension, alignment with human reviewers is 
measured using mean squared error (MSE) and mean absolute error (MAE) 
between predicted scores and annotated ground-truth scores. 
Lower values indicate better alignment. 

To assess rationale quality, we compute the Reason Quality Score (RQS), which measures semantic consistency between generated rationales and human-written explanations under a standardized evaluation prompt (Appendix~\ref{appendix:RQS}). Higher RQS indicates better reasoning fidelity.

We further aggregate alignment and rationale quality into a unified 
\textbf{Human-aligned Score (HAS)}. 
Unless otherwise specified, we set 
$\alpha_1=0.4$, $\alpha_2=0.6$ for combining MSE and MAE into 
an alignment score, and 
$\beta_1=0.75$, $\beta_2=0.25$ for combining alignment and RQS into HAS. 
Lower MSE/MAE and higher RQS/HAS indicate better performance.

\paragraph{Implementation Details.}
Our proposed evaluator, \textbf{SurveyAlign}, is built upon 
Qwen3-32B~\cite{yang2025qwen3technicalreport}. 
We apply Low-Rank Adaptation (LoRA)~\cite{hu2021loralowrankadaptationlarge} 
for parameter-efficient supervised fine-tuning. 
Dimension-specific models are trained following the framework 
described in Section~\ref{sec:benchmark_methods}. 
Detailed hyperparameters and training configurations 
are provided in Appendix~\ref{appendix:sft}.

\paragraph{Input Format.}
All evaluated methods operate under a unified prompting protocol 
(Appendix~\ref{appendix:promptllm}). 
For zero-shot LLM-as-a-Judge baselines and our fine-tuned models, 
we instantiate a standardized prompt template consisting of 
(1) the evaluation dimension definition, 
(2) scoring rubric, 
(3) explicit output format specification, and 
(4) the full survey text. 

For knowledge-augmented settings, the corresponding 
dimension-specific Knowledge Augmentation (KA) representation 
is prepended to the survey text without modifying the remaining 
prompt structure.

For prior evaluators, we follow their original prompting strategies 
and pipeline designs while adapting their outputs to the standardized 
score–rationale format required by our benchmark.

\paragraph{Baselines.}
We compare SurveyAlign against two categories of baselines:

\textbf{(i) Prompt-based LLM-as-a-Judge.}
Off-the-shelf instruction-tuned LLMs are used as zero-shot evaluators 
without task-specific training on SurveyReview. 
For each survey paper, models are executed independently for 
each evaluation dimension, conditioned on the corresponding 
dimension-specific rubric. 
We evaluate GPT-5.2~\cite{openai2025gpt52}, 
Claude-Opus-4.5-Thinking~\cite{anthropic2025claudeopus45}, 
Gemini-3-Pro-Preview~\cite{googledeepmind2025gemini3pro}, 
DeepSeek-v3.2~\cite{deepseekai2025deepseekv32pushingfrontieropen}, 
GLM-4.7~\cite{zai2025glm47}, 
and Qwen3-32B~\cite{yang2025qwen3technicalreport} 
in their zero-shot settings.

\textbf{(ii) Prior Survey Evaluators.}
We additionally compare with LLMMapReduce~\cite{wang2025llmmapreduce}, 
DR-Bench~\cite{du2025deepresearchbenchcomprehensivebenchmark}, 
and SurveyX~\cite{liang2025surveyx}. 
These systems are evaluated using their original methodologies, 
with outputs converted to the standardized score–rationale format 
for fair comparison under our benchmark.

\subsection{Main Results}
\label{subsec:main_results}

\paragraph{Overall alignment.}
Table~\ref{tab:main_results} reports the alignment performance of all evaluators on SurveyReview. 
Our fine-tuned evaluator, \textbf{SurveyAlign} (Qwen3-32B-LoRA), achieves the best performance across all four dimensions. 
It obtains the lowest MSE and MAE on every metric and achieves the highest overall HAS score of 0.74, substantially outperforming all zero-shot baselines and prior work.

\paragraph{Comparison with zero-shot LLM judges.}
Even the strongest zero-shot model (GPT-5.2) reaches an HAS of 0.68, whereas SurveyAlign achieves 0.74. 
Beyond the overall score, SurveyAlign consistently attains lower MSE and MAE on every dimension. This consistent margin indicates that task-specific supervision is necessary to approach reviewer-level consistency.

\paragraph{Dimension-wise improvements.}
Performance gains are observed across both surface-level and knowledge-intensive dimensions. 
Readability and Structure, which rely primarily on text-internal signals, benefit substantially from supervised fine-tuning: 
Readability MSE decreases from 2.13 (GPT-5.2 zero-shot) to 1.43, and Structure from 2.98 to 1.29. 
The particularly large reduction on Structure suggests that document-level organization requires task-specific calibration.

For the knowledge-intensive dimensions—Criticalness and Comprehensiveness—incorporating structured external knowledge further improves alignment. 
Criticalness MSE decreases to 1.52 and Comprehensiveness to 1.26, surpassing all zero-shot baselines. 
These results highlight the importance of explicit knowledge modeling for coverage reasoning and analytical depth.

\paragraph{Rationale quality.}
Beyond numeric scores, we evaluate reasoning quality using RQS. 
While Claude-Opus-4.5-Thinking achieves the highest zero-shot RQS (0.48), it underperforms in score alignment. 
In contrast, SurveyAlign maintains competitive rationale quality (0.36) while delivering substantially better quantitative alignment, demonstrating a favorable trade-off between explanation quality and scoring accuracy.

\paragraph{Comparison with prior evaluators.}
We compare SurveyAlign with representative prior evaluators, including DR-Bench, LLMMapReduce, and SurveyX. 
As these methods were not designed for our multi-dimensional score-plus-rationale setting, we treat them as informative reference points rather than fully matched competitors, reusing their original prompts or formulations whenever feasible. 
The results suggest that while adapted prompting-based evaluators provide useful signals, SurveyAlign achieves more stable reviewer alignment across all four dimensions through direct supervision from human-annotated reviewer data.

\paragraph{Takeaway.}
Overall, results demonstrate that supervised alignment is crucial for reliable survey evaluation, and that knowledge augmentation further strengthens performance on knowledge-intensive criteria.

\subsection{Ablation Study of SurveyAlign}
\label{subsec:ablation}

\subsubsection{Comprehensiveness}

We further examine the effect of introducing Comprehensiveness-specific Knowledge Augmentation (Comp-KA) on the Comprehensiveness dimension under identical training settings.
As shown in Table~\ref{tab:ablation_comp}, incorporating Comp-KA 
reduces MSE from 1.36 to 1.26 and MAE from 0.60 to 0.56.
The consistent improvement across both metrics indicates that 
structured bibliographic context helps the model better assess 
survey coverage and completeness.

\begin{table}[h]
\centering
\caption{Ablation of the SurveyAlign-Comp submodel on the Comprehensiveness (Comp) dimension.}
\label{tab:ablation_comp}
\begin{tabular}{lcc}
\toprule
\textbf{Setting (SurveyAlign-Comp)} & \textbf{MSE} $\downarrow$ & \textbf{MAE} $\downarrow$ \\
\midrule
SFT & 1.36 & 0.60 \\
SFT + Comp-KA & \textbf{1.26} & \textbf{0.56} \\
\bottomrule
\end{tabular}
\end{table}

\subsubsection{Criticalness}

We further examine the interaction between Criticalness-specific Knowledge Augmentation (Crit-KA) and Direct Preference Optimization (DPO) 
on the Criticalness dimension (Table~\ref{tab:ablation_crit}). 
Crit-KA alone does not improve performance and 
slightly increases error, while DPO alone yields only marginal gains. 
In contrast, combining Crit-KA with DPO substantially 
reduces MSE (1.52) and MAE (0.82), achieving the best performance. 
These results suggest that Criticalness evaluation requires both 
access to relevant external knowledge and alignment of reasoning behavior, 
highlighting the complementary roles of dimension-specific 
Knowledge Augmentation and preference optimization.

\begin{table}[h]
\centering
\caption{Ablation of the SurveyAlign-Crit submodel on the Criticalness (Crit.) dimension.}
\label{tab:ablation_crit}
\begin{tabular}{lcc}
\toprule
\textbf{Setting (SurveyAlign-Crit)} & \textbf{MSE} $\downarrow$ & \textbf{MAE} $\downarrow$ \\
\midrule
SFT & 1.96 & 0.93 \\
SFT + Crit-KA & 2.04 & 1.02 \\
SFT + DPO & 1.90 & 0.99 \\
SFT + Crit-KA + DPO & \textbf{1.52} & \textbf{0.82} \\
\bottomrule
\end{tabular}
\end{table}



%% file: Sections/Conclusion.tex
We presented SurveyReview, a reviewer-aligned, multi-dimensional benchmark for evaluating survey papers.
By structuring peer-review reports into four quality dimensions with rationales, the framework enables quantitative measurement of alignment between automatic evaluators and human reviewers.

Experiments show that task-specific supervision substantially improves alignment compared to zero-shot judging, especially on subjective dimensions such as Criticalness and Structure.
Dimension-wise and ablation analyses further reveal distinct roles of knowledge augmentation, preference optimization, and inference strategies.

The benchmark provides standardized splits, evaluation metrics, and diagnostic signals for studying reviewer-aligned assessment.

We hope this benchmark facilitates more transparent and measurable progress in automated survey evaluation.

%% file: Sections/Ethical.tex
This work follows standard ethical guidelines for responsible computing research. The SurveyReview benchmark is built exclusively from publicly available survey papers and open peer review materials originating from OpenReview, F1000Research, and MOPRD. We do not claim ownership of any third‑party texts. We follow the licensing terms, terms of service, and access controls of each source. Consistent with these requirements, we will maximize redistribution of materials that are explicitly redistributable, and for materials that are not redistributable we will provide stable identifiers, provenance, and retrieval instructions so that users can obtain them directly from the original sources under the applicable terms. For metadata and structured annotations that are legally redistributable, we will make them publicly available to the greatest extent permitted, accompanied by clear documentation to facilitate reproducibility and further research.

%% file: Sections/Appendix.tex
\section{Annotation Details}
\label{appendix:annotationdetails}

Annotators label each (survey paper, review report, dimension) unit with a
dimension-specific score and its supporting rationale. The initial annotation
space is $\{-3,-2,-1,0,+1,+2\}$, where $0$ denotes that the dimension is not
mentioned and $-3$ denotes unjudgeable or ambiguous sentiment. These two
special cases are excluded from supervised training and evaluation, leaving
the final benchmark score space $\{-2,-1,+1,+2\}$.

The scoring rubric distinguishes strong negative ($-2$), negative ($-1$),
positive ($+1$), and strong positive ($+2$) assessments. The detailed criteria for each score are specified separately for each evaluation dimension. For each nonzero
dimension, annotators extract the most relevant sentence or sentences from the
review text as rationales. Rationales are kept complete enough to preserve
their original meaning; if several supporting statements are present, they are
recorded as separate rationale snippets.

\subsection{Dimension Definitions}
\label{appendix:dimensiondefinition}

\paragraph{Comprehensiveness}
Evaluates the completeness and relevance of the manuscript’s references. Annotators assess whether key works are cited, important literature is omitted, or outdated/irrelevant sources are used.

\paragraph{Criticalness}
Evaluates the originality and depth of the survey, focusing on whether it provides new insights, critical perspectives, or meaningful synthesis beyond summarizing prior work.  
Related keywords: \textit{insight}, \textit{novel perspective}, \textit{depth}.

\paragraph{Readability}
Evaluates the overall clarity and presentation quality of the manuscript, including whether the writing is clear, well-explained, easy to follow, or difficult to understand.  
Related keywords: \textit{writing}, \textit{presentation}, \textit{easy to follow}, \textit{hard to follow}.

\paragraph{Structure}
Evaluates the logical organization and coherence of sections and subsections, including whether ideas are well-structured and transitions between parts are smooth and logical.

\subsection{Annotation Reliability}
\label{appendix:annotation_reliability}

To assess annotation reliability, we randomly sampled 280 annotation units,
with 70 units for each dimension, and had each unit independently labeled by
two annotators. Overall Krippendorff's $\alpha$ is 0.74. Per-dimension
agreement statistics are reported in Table~\ref{tab:annotation_reliability}.

\begin{table}[H]
\centering
\caption{Annotation reliability statistics by dimension.}
\label{tab:annotation_reliability}
\small
\begin{tabular}{lcccc}
\toprule
\textbf{Dimension} & \textbf{Units} & \textbf{Disagr.} & \textbf{Rate} & \textbf{KP-$\alpha$} \\
\midrule
Comprehensiveness & 70  & 15 & 21.43\% & 0.7702 \\
Criticalness      & 70  & 20 & 28.57\% & 0.7867 \\
Readability       & 70  & 18 & 25.71\% & 0.7260 \\
Structure         & 70  & 17 & 24.29\% & 0.6329 \\
\midrule
Overall           & 280 & 70 & 25.00\% & 0.7400 \\
\bottomrule
\end{tabular}
\end{table}

Among the 70 disagreement cases, 48 differ by one score level and 22 differ
by two score levels, indicating that most remaining disagreements are local
rather than systematic.

\section{Dataset Details}
\label{appendix:datasetdetails}

Table~\ref{tab:dimension_stats} summarizes the dimension-wise score
distribution and rationale density in \textsc{SurveyReview}. Avg.~R denotes
the average number of rationale snippets per annotated instance.

\begin{table}[H]
\centering
\caption{Dimension-wise score and rationale statistics.}
\label{tab:dimension_stats}
\small
\begin{tabular}{lccc}
\toprule
\textbf{Dimension} & \textbf{Mean Score} & \textbf{Std} & \textbf{Avg. R} \\
\midrule
Readability & $-0.36$ & $0.94$ & 3.54 \\
Structure & $-0.40$ & $0.83$ & 3.07 \\
Comprehensiveness & $-0.29$ & $0.90$ & 2.84 \\
Criticalness & $-0.17$ & $0.98$ & 2.80 \\
\bottomrule
\end{tabular}
\end{table}

Mean scores range from $-0.40$ for Structure to $-0.17$ for Criticalness,
showing a moderate skew toward negative assessments across dimensions. The
standard deviations are close to 1.0, indicating substantial variation in
reviewer judgments. Avg.~R ranges from 2.80 to 3.54, suggesting that reviewers
often provide multiple rationale snippets for each annotated dimension.
Statistics are computed at the (paper, review, dimension) level because one
survey paper may receive multiple review reports.


\section{Evaluation Prompting Protocol}
\label{appendix:promptllm}

All evaluated methods are instantiated under a unified dimension-specific
prompting protocol. This protocol is used for zero-shot LLM-as-a-judge
baselines, for constructing supervised fine-tuning examples, and for
knowledge-augmented evaluation. Each prompt contains four components:
(1) a reviewer role instruction, (2) the target evaluation dimension
definition, (3) dimension-specific scoring criteria, and (4) an explicit
score--rationale output format. The valid score set is
$\{-2,-1,+1,+2\}$.

\subsection{Unified Prompt Template}

\begin{verbatim}
Please act as a research paper reviewer and evaluate the
following survey from the perspective of [Target Dimension].

[Knowledge Augmentation Description, if applicable]

[Definition of Target Dimension]

[Dimension-Specific Scoring Criteria]

### Output Requirements:
Provide:
(1) A brief rationale explaining your evaluation.
(2) A numeric score chosen from {-2, -1, 1, 2}.

### Output Format:
Reason$$$X$$$

----------------------------
[Knowledge Augmentation Content, if applicable]

Survey to Evaluate:
{FullSurveyText}
----------------------------
\end{verbatim}

\paragraph{Use in zero-shot evaluation.}
For prompt-based LLM-as-a-judge baselines, the knowledge augmentation fields
are omitted. The model receives the dimension definition, the corresponding
scoring criteria, and the full survey text, and then produces a rationale and
a score in the required format.

\paragraph{Use in supervised fine-tuning.}
The same schema is used to construct instruction-tuning examples. The
\texttt{instruction} field contains the dimension-specific reviewer
instruction and scoring criteria, the \texttt{input} field contains the survey
text, and the \texttt{output} field contains the human-annotated rationale and
numeric score. This keeps supervised fine-tuning aligned with the zero-shot
evaluation setting.

\paragraph{Use with knowledge augmentation.}
For knowledge-intensive dimensions, the same prompt structure is retained, but
a dimension-specific external context block is prepended before the survey
text. For \textit{Comprehensiveness}, this block contains the original
bibliography and an expanded reference set for literature-coverage comparison.
For \textit{Criticalness}, it contains chronological summaries of related
expanded works to support assessment of analytical depth and synthesis. The
output format remains unchanged across all settings, ensuring comparable
score--rationale outputs.

\section{SFT Details}
\label{appendix:sft}

We fine-tune separate Qwen3-32B models for each evaluation dimension using
LoRA. The LoRA rank is $r=8$, the scaling factor is $\alpha=16$, and dropout
is set to 0.0. LoRA is applied to all attention and feed-forward projection
layers, including \texttt{q\_proj}, \texttt{k\_proj}, \texttt{v\_proj},
\texttt{o\_proj}, \texttt{gate\_proj}, \texttt{up\_proj}, and
\texttt{down\_proj}; the base model weights remain frozen. Training uses
AdamW with learning rate $5\times10^{-5}$, a cosine learning rate scheduler,
20 warmup steps, 15 epochs, per-device batch size 1, gradient accumulation
steps of 16, and a maximum sequence length of 16{,}384 tokens. We train in
fp16 precision with gradient checkpointing, DeepSpeed ZeRO-3, and
FlashAttention-2. The best checkpoint is selected by validation loss under a
90\% / 10\% train--validation split.


\section{Test-Time Scaling Details}
\label{appendix:testtimescaling}

For each evaluation dimension, we sample $K=5$ independent outputs with
temperature $T=0.7$. The final score is selected by majority voting:
\[
s_{\mathrm{final}} =
\arg\max_{s} \sum_{i=1}^{K} \mathbb{1}[s_i=s].
\]
In the case of ties, we choose the score with the highest average generation
log-probability. The rationales from all runs are then synthesized into a
single explanation conditioned on the dimension rubric and the consensus
score, ensuring that the final rationale is consistent with
$s_{\mathrm{final}}$.

\section{Knowledge Augmentation Details}
\label{appendix:ka_details}

This appendix describes the construction of the 
dimension-specific Knowledge Augmentation (KA) 
content used for knowledge-intensive dimensions.

\subsection{Comprehensiveness-specific Knowledge Augmentation (Comp-KA)}
\label{appendix:Comp-KA}

Let $R$ denote the reference set cited by a survey, and $R^{+}$ denote the
expanded citation neighborhood constructed by the toolkit.

\paragraph{Connectivity scoring.}
For each candidate paper $p \in R^{+}$, we define its connectivity to $R$ as:
\[
\text{score}(p) = \sum_{r \in R} \mathbb{1}[\, p \leftrightarrow r \,],
\]
where $(p \leftrightarrow r)$ indicates that $p$ and $r$ have a citation
relationship (either direction).

\paragraph{Selection and serialization.}
Given $n = |R|$, we select the top-$2n$ papers in $R^{+}$ ranked by
$\text{score}(p)$ and combine them with the original set $R$:
\[
\text{Comp-KA} = R \cup \operatorname{Top}_{2n}(R^{+}).
\]
The resulting Comp-KA content is serialized as a list of paper entries,
each containing title, publication year, and citation count, 
providing structured signals for coverage assessment.

\subsection{Criticalness-specific Knowledge Augmentation (Crit-KA)}
\label{appendix:Crit-KA}

We construct a temporally structured representation from $R^{+}$ 
to support analysis-oriented evaluation.

\paragraph{Thresholding and candidate set.}
We compute adaptive thresholds from the original reference set $R$:
\[
\tau_c = \text{Percentile}_\alpha(\{\text{citations}(r)\}_{r \in R}), 
\quad
\tau_y = \text{Percentile}_\beta(\{\text{year}(r)\}_{r \in R}),
\]
where $\alpha$ and $\beta$ are fixed hyperparameters.

We then form a candidate set:
\[
C = \{\, p \in R^{+} \mid 
\text{citations}(p) \ge \tau_c 
\;\lor\; 
\text{year}(p) \ge \tau_y \,\}.
\]
Intuitively, $C$ includes papers that are either highly cited 
or relatively recent compared with the survey's references.

\paragraph{Stage 1: Temporal block summarization under a token budget.}
We sort $C$ by publication year in ascending order.
Under a predefined token budget $B$, we sequentially pack 
(title, abstract) pairs into a buffer until reaching the budget, 
forming a temporal block.

For each block, we record its year interval
$\Delta t = [y_{\min}, y_{\max}]$ and apply an LLM to produce 
a concise block summary:
\[
(\Delta t_i, \text{Summary}_i) 
= \text{Summarize}(\text{Block}_i).
\]

\paragraph{Stage 2: Global compression to a fixed length.}
Given the set of timestamped block summaries 
$\{(\Delta t_i, \text{Summary}_i)\}$, 
we perform a second-stage compression to obtain a final representation 
of fixed length $L$:
\[
\text{Crit-KA} 
= \text{Compress}(\{(\Delta t_i, \text{Summary}_i)\}, L).
\]

The output preserves coarse temporal structure through $\Delta t_i$ 
while enforcing a global length constraint.


\section{Reason Quality Score}
\label{appendix:RQS}

We compute the Reason Quality Score (RQS) using GPT-5.2 as a meta-evaluator
to assess semantic consistency between predicted rationales and human-written
reference rationales. For each instance, the judge receives the evaluation
dimension, paper title, reference rationale, and predicted rationale, and
assigns one score from $\{0.0,0.2,0.4,0.6,0.8,1.0\}$, where higher values
indicate stronger rationale alignment. RQS is computed by averaging these
scores across samples and then across dimensions.

To validate the judge, we also conduct a small human meta-evaluation on 50
rationale pairs. The GPT-5.2 judge scores show strong agreement with the mean
human ratings (Pearson $r=0.804$, Spearman $\rho=0.795$), suggesting that RQS
provides a meaningful soft signal for rationale quality.



\section{Baseline Details}
\label{appendix:baselines}

For prompt-based LLM-as-a-judge baselines, we use the same unified
dimension-specific prompt format described in Appendix~\ref{appendix:promptllm}.
For prior survey evaluators, we follow their original evaluation designs when
available and adapt their outputs to the SurveyReview score--rationale format.
When a method uses a different scoring scale, we map its output to the closest
category in our unified score space $\{-2,-1,+1,+2\}$.

\paragraph{LLMMapReduce-V2.}
We follow its original prompt design and use its recommended model setting for
long-form survey evaluation.

\paragraph{DeepResearch-Bench.}
Because SurveyReview does not provide gold-standard reference survey answers,
we exclude reference-answer-dependent components and align its insight score
with our Criticalness dimension.

\paragraph{SurveyX}
We follow the evaluation and prompting configuration described in SurveyX and
convert its outputs into our standardized score--rationale format.



\section{Additional Robustness and Diagnostic Analyses}
\label{appendix:additinal_rob}

\subsection{Source-wise Evaluation}
\label{appendix:source_wise_eval}

Table~\ref{tab:source_wise_has} reports source-wise HAS on F1000Research,
MOPRD, and OpenReview. SurveyAlign ranks first on all three sources, showing
that the main conclusion is stable across data sources.

\begin{table}[H]
\centering
\footnotesize
\setlength{\tabcolsep}{4.0pt}
\caption{Source-wise HAS comparison across benchmark sources.}
\label{tab:source_wise_has}
\begin{tabular}{lccc}
\toprule
\textbf{Model} & \textbf{F1000} & \textbf{MOPRD} & \textbf{OpenReview} \\
\midrule
SurveyAlign       & \textbf{0.75} & \textbf{0.71} & \textbf{0.66} \\
GPT-5.2           & 0.70 & 0.62 & 0.65 \\
Gemini-3-pro      & 0.60 & 0.47 & 0.54 \\
DeepSeek-v3.2     & 0.61 & 0.53 & 0.56 \\
GLM-4.7           & 0.63 & 0.50 & 0.56 \\
Claude-Opus-4.5   & 0.71 & 0.57 & 0.59 \\
\bottomrule
\end{tabular}
\end{table}

The absolute HAS values vary across sources: F1000Research appears easier,
whereas MOPRD and OpenReview are more challenging. However, the ranking is
stable across all sources. SurveyAlign obtains the highest HAS on
F1000Research, MOPRD, and OpenReview, indicating that source composition
changes the metric scale but does not change the main comparative conclusion.

\subsection{Meta-Evaluation of the LLM Judge}
\label{app:judge_meta_eval}

We validate the GPT-5.2 rationale judge with a 50-pair human
meta-evaluation. Three human annotators independently assess rationale
consistency, yielding human-human agreement of Pearson $r=0.705$. The GPT-5.2
judge scores show strong agreement with mean human ratings (Pearson
$r=0.804$, Spearman $\rho=0.795$), supporting RQS as a meaningful soft signal.

\subsection{Sensitivity Analysis of HAS Weighting}
\label{appendix:has_weighting}

Table~\ref{tab:has_weighting} varies the RQS weight in HAS. SurveyAlign remains
ranked first under the default setting ($\beta_{\mathrm{RQS}}=0.25$) and the
nearby moderate setting ($\beta_{\mathrm{RQS}}=0.40$).

\begin{table}[H]
\centering
\scriptsize
\setlength{\tabcolsep}{2.2pt}
\caption{HAS sensitivity under different RQS weights.}
\label{tab:has_weighting}
\begin{tabular}{lcccccc}
\toprule
\textbf{Model} & \textbf{Align} & \textbf{RQS}
& \textbf{0.25} & \textbf{0.40} & \textbf{0.60} & \textbf{0.75} \\
\midrule
SurveyAlign       & 0.86 & 0.36 & 0.74 & 0.66 & 0.56 & 0.49 \\
GPT-5.2           & 0.77 & 0.42 & 0.68 & 0.63 & 0.56 & 0.51 \\
Gemini-3-pro      & 0.67 & 0.29 & 0.58 & 0.52 & 0.44 & 0.39 \\
DeepSeek-v3.2     & 0.65 & 0.37 & 0.58 & 0.54 & 0.48 & 0.44 \\
GLM-4.7           & 0.68 & 0.37 & 0.60 & 0.55 & 0.49 & 0.45 \\
Claude-Opus-4.5   & 0.74 & 0.48 & 0.68 & 0.64 & 0.59 & 0.55 \\
\bottomrule
\end{tabular}
\end{table}

These results show that the default HAS setting is not overly sensitive to a
single weighting choice. SurveyAlign remains first when
$\beta_{\mathrm{RQS}}=0.25$ and $0.40$. When rationale quality receives a much
larger weight, models with stronger raw RQS become more competitive; this is
expected because SurveyAlign's main advantage is human-grounded score
alignment. Thus, the intended benchmark setting and nearby alternatives
support the same main conclusion.

\subsection{Multi-Judge Evaluation for RQS and HAS}
\label{appendix:multi_judge_eval}

To test whether the conclusion depends on a single rationale judge, we
recompute RQS and HAS with GPT-5.2, Claude-Opus-4.5, and Gemini-3-pro as
judges. Although RQS values vary by judge, SurveyAlign achieves the highest
HAS under all three judge choices.

\begin{table}[H]
\centering
\footnotesize
\setlength{\tabcolsep}{4.0pt}
\caption{Multi-judge RQS results.}
\label{tab:multi_judge_rqs}
\begin{tabular}{lccc}
\toprule
\textbf{Model} & \textbf{GPT-5.2} & \textbf{Claude} & \textbf{Gemini} \\
\midrule
SurveyAlign       & 0.36 & 0.41 & 0.22 \\
GPT-5.2           & 0.42 & 0.47 & 0.31 \\
Gemini-3-pro      & 0.29 & 0.46 & 0.31 \\
DeepSeek-v3.2     & 0.38 & 0.41 & 0.28 \\
GLM-4.7           & 0.37 & 0.44 & 0.30 \\
Claude-Opus-4.5   & \textbf{0.48} & \textbf{0.50} & \textbf{0.39} \\
\bottomrule
\end{tabular}
\end{table}

\begin{table}[H]
\centering
\footnotesize
\setlength{\tabcolsep}{4.0pt}
\caption{Multi-judge HAS results after recomputing RQS.}
\label{tab:multi_judge_has}
\begin{tabular}{lccc}
\toprule
\textbf{Model} & \textbf{GPT-5.2} & \textbf{Claude} & \textbf{Gemini} \\
\midrule
SurveyAlign       & \textbf{0.74} & \textbf{0.75} & \textbf{0.71} \\
GPT-5.2           & 0.68 & 0.69 & 0.65 \\
Gemini-3-pro      & 0.58 & 0.62 & 0.59 \\
DeepSeek-v3.2     & 0.58 & 0.59 & 0.56 \\
GLM-4.7           & 0.60 & 0.62 & 0.58 \\
Claude-Opus-4.5   & 0.68 & 0.68 & 0.66 \\
\bottomrule
\end{tabular}
\end{table}

The RQS values differ across judges, confirming that rationale-quality scoring
is judge-sensitive. Nevertheless, after recomputing HAS with each judge,
SurveyAlign remains the top-ranked method under GPT-5.2, Claude-Opus-4.5, and
Gemini-3-pro. This suggests that the final system-level conclusion is robust
even though the rationale component itself should be interpreted as a softer
diagnostic signal.

%% file: main.bbl

\begin{thebibliography}{37}


\ifx \showCODEN    \undefined \def \showCODEN     #1{\unskip}     \fi
\ifx \showISBNx    \undefined \def \showISBNx     #1{\unskip}     \fi
\ifx \showISBNxiii \undefined \def \showISBNxiii  #1{\unskip}     \fi
\ifx \showISSN     \undefined \def \showISSN      #1{\unskip}     \fi
\ifx \showLCCN     \undefined \def \showLCCN      #1{\unskip}     \fi
\ifx \shownote     \undefined \def \shownote      #1{#1}          \fi
\ifx \showarticletitle \undefined \def \showarticletitle #1{#1}   \fi
\ifx \showURL      \undefined \def \showURL       {\relax}        \fi
\providecommand\bibfield[2]{#2}
\providecommand\bibinfo[2]{#2}
\providecommand\natexlab[1]{#1}
\providecommand\showeprint[2][]{arXiv:#2}

\bibitem[gro(2026)]%
        {grobid}
 \bibinfo{year}{2008--2026}\natexlab{}.
\newblock \bibinfo{title}{GROBID}.
\newblock \bibinfo{howpublished}{\url{https://github.com/kermitt2/grobid}}.
\newblock
\showeprint[swh]{1:dir:dab86b296e3c3216e2241968f0d63b68e8209d3c}


\bibitem[Agarwal et~al\mbox{.}(2025)]%
        {agarwal2025artscalingtesttimecompute}
\bibfield{author}{\bibinfo{person}{Aradhye Agarwal}, \bibinfo{person}{Ayan
  Sengupta}, {and} \bibinfo{person}{Tanmoy Chakraborty}.}
  \bibinfo{year}{2025}\natexlab{}.
\newblock \bibinfo{title}{The Art of Scaling Test-Time Compute for Large
  Language Models}.
\newblock
\showeprint[arxiv]{2512.02008}~[cs.CL]
\urldef\tempurl%
\url{https://arxiv.org/abs/2512.02008}
\showURL{%
\tempurl}


\bibitem[{Anthropic}(2025)]%
        {anthropic2025claudeopus45}
\bibfield{author}{\bibinfo{person}{{Anthropic}}.}
  \bibinfo{year}{2025}\natexlab{}.
\newblock \bibinfo{title}{Introducing Claude Opus 4.5}.
\newblock
\urldef\tempurl%
\url{https://www.anthropic.com/news/claude-opus-4-5}
\showURL{%
\tempurl}


\bibitem[Chen et~al\mbox{.}(2025)]%
        {chen2025surveygen}
\bibfield{author}{\bibinfo{person}{Jing Chen}, \bibinfo{person}{Zhiheng Yang},
  \bibinfo{person}{Yixian Shen}, \bibinfo{person}{Jie Liu},
  \bibinfo{person}{Adam Belloum}, \bibinfo{person}{Chrysa Papagainni}, {and}
  \bibinfo{person}{Paola Grosso}.} \bibinfo{year}{2025}\natexlab{}.
\newblock \bibinfo{title}{SurveyGen-I: Consistent Scientific Survey Generation
  with Evolving Plans and Memory-Guided Writing}.
\newblock
\showeprint[arxiv]{2508.14317}~[cs.CL]
\urldef\tempurl%
\url{https://arxiv.org/abs/2508.14317}
\showURL{%
\tempurl}


\bibitem[{DeepSeek-AI}(2025)]%
        {deepseekai2025deepseekv32pushingfrontieropen}
\bibfield{author}{\bibinfo{person}{{DeepSeek-AI}}.}
  \bibinfo{year}{2025}\natexlab{}.
\newblock \bibinfo{title}{DeepSeek-V3.2: Pushing the Frontier of Open Large
  Language Models}.
\newblock
\showeprint[arxiv]{2512.02556}~[cs.CL]
\urldef\tempurl%
\url{https://arxiv.org/abs/2512.02556}
\showURL{%
\tempurl}


\bibitem[Du et~al\mbox{.}(2025)]%
        {du2025deepresearchbenchcomprehensivebenchmark}
\bibfield{author}{\bibinfo{person}{Mingxuan Du}, \bibinfo{person}{Benfeng Xu},
  \bibinfo{person}{Chiwei Zhu}, \bibinfo{person}{Xiaorui Wang}, {and}
  \bibinfo{person}{Zhendong Mao}.} \bibinfo{year}{2025}\natexlab{}.
\newblock \bibinfo{title}{DeepResearch Bench: A Comprehensive Benchmark for
  Deep Research Agents}.
\newblock
\showeprint[arxiv]{2506.11763}~[cs.CL]
\urldef\tempurl%
\url{https://arxiv.org/abs/2506.11763}
\showURL{%
\tempurl}


\bibitem[Feng et~al\mbox{.}(2025)]%
        {feng2025sampleefficienthumanevaluationlarge}
\bibfield{author}{\bibinfo{person}{Kehua Feng}, \bibinfo{person}{Keyan Ding},
  \bibinfo{person}{Hongzhi Tan}, \bibinfo{person}{Kede Ma},
  \bibinfo{person}{Zhihua Wang}, \bibinfo{person}{Shuangquan Guo},
  \bibinfo{person}{Yuzhou Cheng}, \bibinfo{person}{Ge Sun},
  \bibinfo{person}{Guozhou Zheng}, \bibinfo{person}{Qiang Zhang}, {and}
  \bibinfo{person}{Huajun Chen}.} \bibinfo{year}{2025}\natexlab{}.
\newblock \bibinfo{title}{Sample-Efficient Human Evaluation of Large Language
  Models via Maximum Discrepancy Competition}.
\newblock
\showeprint[arxiv]{2404.08008}~[cs.LG]
\urldef\tempurl%
\url{https://arxiv.org/abs/2404.08008}
\showURL{%
\tempurl}


\bibitem[{Google DeepMind}(2025)]%
        {googledeepmind2025gemini3pro}
\bibfield{author}{\bibinfo{person}{{Google DeepMind}}.}
  \bibinfo{year}{2025}\natexlab{}.
\newblock \bibinfo{title}{Gemini 3 Pro Model Card}.
\newblock
\urldef\tempurl%
\url{https://deepmind.google/models/model-cards/gemini-3-pro/}
\showURL{%
\tempurl}


\bibitem[Hu et~al\mbox{.}(2021)]%
        {hu2021loralowrankadaptationlarge}
\bibfield{author}{\bibinfo{person}{Edward~J. Hu}, \bibinfo{person}{Yelong
  Shen}, \bibinfo{person}{Phillip Wallis}, \bibinfo{person}{Zeyuan Allen-Zhu},
  \bibinfo{person}{Yuanzhi Li}, \bibinfo{person}{Shean Wang},
  \bibinfo{person}{Lu Wang}, {and} \bibinfo{person}{Weizhu Chen}.}
  \bibinfo{year}{2021}\natexlab{}.
\newblock \bibinfo{title}{LoRA: Low-Rank Adaptation of Large Language Models}.
\newblock
\showeprint[arxiv]{2106.09685}~[cs.CL]
\urldef\tempurl%
\url{https://arxiv.org/abs/2106.09685}
\showURL{%
\tempurl}


\bibitem[Huotala et~al\mbox{.}(2025)]%
        {huotala2025aisysrevllmbasedtool}
\bibfield{author}{\bibinfo{person}{Aleksi Huotala}, \bibinfo{person}{Miikka
  Kuutila}, \bibinfo{person}{Olli-Pekka Turtio}, {and} \bibinfo{person}{Mika
  Mäntylä}.} \bibinfo{year}{2025}\natexlab{}.
\newblock \bibinfo{title}{AISysRev -- LLM-based Tool for Title-abstract
  Screening}.
\newblock
\showeprint[arxiv]{2510.06708}~[cs.SE]
\urldef\tempurl%
\url{https://arxiv.org/abs/2510.06708}
\showURL{%
\tempurl}


\bibitem[Li et~al\mbox{.}(2025)]%
        {li2025reportbenchevaluatingdeepresearch}
\bibfield{author}{\bibinfo{person}{Minghao Li}, \bibinfo{person}{Ying Zeng},
  \bibinfo{person}{Zhihao Cheng}, \bibinfo{person}{Cong Ma}, {and}
  \bibinfo{person}{Kai Jia}.} \bibinfo{year}{2025}\natexlab{}.
\newblock \bibinfo{title}{ReportBench: Evaluating Deep Research Agents via
  Academic Survey Tasks}.
\newblock
\showeprint[arxiv]{2508.15804}~[cs.CL]
\urldef\tempurl%
\url{https://arxiv.org/abs/2508.15804}
\showURL{%
\tempurl}


\bibitem[Liang et~al\mbox{.}(2025)]%
        {liang2025surveyx}
\bibfield{author}{\bibinfo{person}{Xun Liang}, \bibinfo{person}{Jiawei Yang},
  \bibinfo{person}{Yezhaohui Wang}, \bibinfo{person}{Chen Tang},
  \bibinfo{person}{Zifan Zheng}, \bibinfo{person}{Shichao Song},
  \bibinfo{person}{Zehao Lin}, \bibinfo{person}{Yebin Yang},
  \bibinfo{person}{Simin Niu}, \bibinfo{person}{Hanyu Wang},
  \bibinfo{person}{Bo Tang}, \bibinfo{person}{Feiyu Xiong},
  \bibinfo{person}{Keming Mao}, {and} \bibinfo{person}{Zhiyu li}.}
  \bibinfo{year}{2025}\natexlab{}.
\newblock \bibinfo{title}{SurveyX: Academic Survey Automation via Large
  Language Models}.
\newblock
\showeprint[arxiv]{2502.14776}~[cs.CL]
\urldef\tempurl%
\url{https://arxiv.org/abs/2502.14776}
\showURL{%
\tempurl}


\bibitem[Lin(2004)]%
        {lin-2004-rouge}
\bibfield{author}{\bibinfo{person}{Chin-Yew Lin}.}
  \bibinfo{year}{2004}\natexlab{}.
\newblock \showarticletitle{{ROUGE}: A Package for Automatic Evaluation of
  Summaries}. In \bibinfo{booktitle}{\emph{Text Summarization Branches Out}}.
  \bibinfo{publisher}{Association for Computational Linguistics},
  \bibinfo{address}{Barcelona, Spain}, \bibinfo{pages}{74--81}.
\newblock
\urldef\tempurl%
\url{https://aclanthology.org/W04-1013/}
\showURL{%
\tempurl}


\bibitem[Lin et~al\mbox{.}(2023)]%
        {lin2023moprd}
\bibfield{author}{\bibinfo{person}{Jialiang Lin}, \bibinfo{person}{Jiaxin
  Song}, \bibinfo{person}{Zhangping Zhou}, \bibinfo{person}{Yidong Chen}, {and}
  \bibinfo{person}{Xiaodong Shi}.} \bibinfo{year}{2023}\natexlab{}.
\newblock \showarticletitle{MOPRD: A multidisciplinary open peer review
  dataset}.
\newblock \bibinfo{journal}{\emph{Neural Computing and Applications}}
  \bibinfo{volume}{35}, \bibinfo{number}{34} (\bibinfo{date}{Sept.}
  \bibinfo{year}{2023}), \bibinfo{pages}{24191–24206}.
\newblock
\showISSN{1433-3058}
\href{https://doi.org/10.1007/s00521-023-08891-5}{doi:\nolinkurl{10.1007/s00521-023-08891-5}}


\bibitem[{OpenAI}(2025)]%
        {openai2025gpt52}
\bibfield{author}{\bibinfo{person}{{OpenAI}}.} \bibinfo{year}{2025}\natexlab{}.
\newblock \bibinfo{title}{Introducing GPT-5.2}.
\newblock
\urldef\tempurl%
\url{https://openai.com/index/introducing-gpt-5-2/}
\showURL{%
\tempurl}


\bibitem[Papineni et~al\mbox{.}(2002)]%
        {2022bleu}
\bibfield{author}{\bibinfo{person}{Kishore Papineni}, \bibinfo{person}{Salim
  Roukos}, \bibinfo{person}{Todd Ward}, {and} \bibinfo{person}{Wei-Jing Zhu}.}
  \bibinfo{year}{2002}\natexlab{}.
\newblock \showarticletitle{BLEU: a method for automatic evaluation of machine
  translation}. In \bibinfo{booktitle}{\emph{Proceedings of the 40th Annual
  Meeting on Association for Computational Linguistics}} (Philadelphia,
  Pennsylvania) \emph{(\bibinfo{series}{ACL '02})}.
  \bibinfo{publisher}{Association for Computational Linguistics},
  \bibinfo{address}{USA}, \bibinfo{pages}{311–318}.
\newblock
\href{https://doi.org/10.3115/1073083.1073135}{doi:\nolinkurl{10.3115/1073083.1073135}}


\bibitem[Paruchuri(2025)]%
        {marker}
\bibfield{author}{\bibinfo{person}{Vik Paruchuri}.}
  \bibinfo{year}{2025}\natexlab{}.
\newblock \bibinfo{title}{Marker}.
\newblock \bibinfo{howpublished}{\url{https://github.com/VikParuchuri/marker}}.
\newblock
\newblock
\shownote{GitHub repository}.


\bibitem[Rafailov et~al\mbox{.}(2024)]%
        {rafailov2024directpreferenceoptimizationlanguage}
\bibfield{author}{\bibinfo{person}{Rafael Rafailov}, \bibinfo{person}{Archit
  Sharma}, \bibinfo{person}{Eric Mitchell}, \bibinfo{person}{Stefano Ermon},
  \bibinfo{person}{Christopher~D. Manning}, {and} \bibinfo{person}{Chelsea
  Finn}.} \bibinfo{year}{2024}\natexlab{}.
\newblock \bibinfo{title}{Direct Preference Optimization: Your Language Model
  is Secretly a Reward Model}.
\newblock
\showeprint[arxiv]{2305.18290}~[cs.LG]
\urldef\tempurl%
\url{https://arxiv.org/abs/2305.18290}
\showURL{%
\tempurl}


\bibitem[Schmidtová et~al\mbox{.}(2024)]%
        {2024automatic}
\bibfield{author}{\bibinfo{person}{Patrícia Schmidtová},
  \bibinfo{person}{Saad Mahamood}, \bibinfo{person}{Simone Balloccu},
  \bibinfo{person}{Ondřej Dušek}, \bibinfo{person}{Albert Gatt},
  \bibinfo{person}{Dimitra Gkatzia}, \bibinfo{person}{David~M. Howcroft},
  \bibinfo{person}{Ondřej Plátek}, {and} \bibinfo{person}{Adarsa
  Sivaprasad}.} \bibinfo{year}{2024}\natexlab{}.
\newblock \bibinfo{title}{Automatic Metrics in Natural Language Generation: A
  Survey of Current Evaluation Practices}.
\newblock
\showeprint[arxiv]{2408.09169}~[cs.CL]
\urldef\tempurl%
\url{https://arxiv.org/abs/2408.09169}
\showURL{%
\tempurl}


\bibitem[Shi et~al\mbox{.}(2025)]%
        {shi2025scisage}
\bibfield{author}{\bibinfo{person}{Xiaofeng Shi}, \bibinfo{person}{Qian Kou},
  \bibinfo{person}{Yuduo Li}, \bibinfo{person}{Ning Tang},
  \bibinfo{person}{Jinxin Xie}, \bibinfo{person}{Longbin Yu},
  \bibinfo{person}{Songjing Wang}, {and} \bibinfo{person}{Hua Zhou}.}
  \bibinfo{year}{2025}\natexlab{}.
\newblock \bibinfo{title}{SciSage: A Multi-Agent Framework for High-Quality
  Scientific Survey Generation}.
\newblock
\showeprint[arxiv]{2506.12689}~[cs.AI]
\urldef\tempurl%
\url{https://arxiv.org/abs/2506.12689}
\showURL{%
\tempurl}


\bibitem[Su et~al\mbox{.}(2026)]%
        {su2026surge}
\bibfield{author}{\bibinfo{person}{Weihang Su}, \bibinfo{person}{Anzhe Xie},
  \bibinfo{person}{Qingyao Ai}, \bibinfo{person}{Jianming Long},
  \bibinfo{person}{Xuanyi Chen}, \bibinfo{person}{Jiaxin Mao},
  \bibinfo{person}{Ziyi Ye}, {and} \bibinfo{person}{Yiqun Liu}.}
  \bibinfo{year}{2026}\natexlab{}.
\newblock \bibinfo{title}{SurGE: A Benchmark and Evaluation Framework for
  Scientific Survey Generation}.
\newblock
\showeprint[arxiv]{2508.15658}~[cs.CL]
\urldef\tempurl%
\url{https://arxiv.org/abs/2508.15658}
\showURL{%
\tempurl}


\bibitem[Sun et~al\mbox{.}(2025a)]%
        {sun2025openreviewprotectedleveragedcommunity}
\bibfield{author}{\bibinfo{person}{Hao Sun}, \bibinfo{person}{Yunyi Shen},
  {and} \bibinfo{person}{Mihaela van~der Schaar}.}
  \bibinfo{year}{2025}\natexlab{a}.
\newblock \bibinfo{title}{OpenReview Should be Protected and Leveraged as a
  Community Asset for Research in the Era of Large Language Models}.
\newblock
\showeprint[arxiv]{2505.21537}~[cs.CY]
\urldef\tempurl%
\url{https://arxiv.org/abs/2505.21537}
\showURL{%
\tempurl}


\bibitem[Sun et~al\mbox{.}(2025b)]%
        {sun2025surveybench}
\bibfield{author}{\bibinfo{person}{Zhaojun Sun}, \bibinfo{person}{Xuzhou Zhu},
  \bibinfo{person}{Xuanhe Zhou}, \bibinfo{person}{Xin Tong},
  \bibinfo{person}{Shuo Wang}, \bibinfo{person}{Jie Fu},
  \bibinfo{person}{Guoliang Li}, \bibinfo{person}{Zhiyuan Liu}, {and}
  \bibinfo{person}{Fan Wu}.} \bibinfo{year}{2025}\natexlab{b}.
\newblock \bibinfo{title}{SurveyBench: Can LLM(-Agents) Write Academic Surveys
  that Align with Reader Needs?}
\newblock
\showeprint[arxiv]{2510.03120}~[cs.CL]
\urldef\tempurl%
\url{https://arxiv.org/abs/2510.03120}
\showURL{%
\tempurl}


\bibitem[Tang(2016)]%
        {tang2016aminer}
\bibfield{author}{\bibinfo{person}{Jie Tang}.} \bibinfo{year}{2016}\natexlab{}.
\newblock \showarticletitle{AMiner: Toward Understanding Big Scholar Data}. In
  \bibinfo{booktitle}{\emph{Proceedings of the Ninth ACM International
  Conference on Web Search and Data Mining}} (San Francisco, California, USA)
  \emph{(\bibinfo{series}{WSDM '16})}. \bibinfo{publisher}{Association for
  Computing Machinery}, \bibinfo{address}{New York, NY, USA},
  \bibinfo{pages}{467}.
\newblock
\showISBNx{9781450337168}
\href{https://doi.org/10.1145/2835776.2835849}{doi:\nolinkurl{10.1145/2835776.2835849}}


\bibitem[Thelwall et~al\mbox{.}(2019)]%
        {thelwall2019doesuseopennonanonymous}
\bibfield{author}{\bibinfo{person}{Mike Thelwall}, \bibinfo{person}{Verena
  Weigert}, \bibinfo{person}{Liz Allen}, \bibinfo{person}{Zena Nyakoojo}, {and}
  \bibinfo{person}{Eleanor-Rose Papas}.} \bibinfo{year}{2019}\natexlab{}.
\newblock \bibinfo{title}{Does the use of open, non-anonymous peer review in
  scholarly publishing introduce bias? Evidence from the F1000 post-publication
  open peer review publishing model}.
\newblock
\showeprint[arxiv]{1911.03379}~[cs.DL]
\urldef\tempurl%
\url{https://arxiv.org/abs/1911.03379}
\showURL{%
\tempurl}


\bibitem[Wang et~al\mbox{.}(2025)]%
        {wang2025llmmapreduce}
\bibfield{author}{\bibinfo{person}{Haoyu Wang}, \bibinfo{person}{Yujia Fu},
  \bibinfo{person}{Zhu Zhang}, \bibinfo{person}{Shuo Wang},
  \bibinfo{person}{Zirui Ren}, \bibinfo{person}{Xiaorong Wang},
  \bibinfo{person}{Zhili Li}, \bibinfo{person}{Chaoqun He}, \bibinfo{person}{Bo
  An}, \bibinfo{person}{Zhiyuan Liu}, {and} \bibinfo{person}{Maosong Sun}.}
  \bibinfo{year}{2025}\natexlab{}.
\newblock \bibinfo{title}{LLM$\times$MapReduce-V2: Entropy-Driven Convolutional
  Test-Time Scaling for Generating Long-Form Articles from Extremely Long
  Resources}.
\newblock
\showeprint[arxiv]{2504.05732}~[cs.CL]
\urldef\tempurl%
\url{https://arxiv.org/abs/2504.05732}
\showURL{%
\tempurl}


\bibitem[Wang et~al\mbox{.}(2024)]%
        {wang2024autosurvey}
\bibfield{author}{\bibinfo{person}{Yidong Wang}, \bibinfo{person}{Qi Guo},
  \bibinfo{person}{Wenjin Yao}, \bibinfo{person}{Hongbo Zhang},
  \bibinfo{person}{Xin Zhang}, \bibinfo{person}{Zhen Wu},
  \bibinfo{person}{Meishan Zhang}, \bibinfo{person}{Xinyu Dai},
  \bibinfo{person}{Min Zhang}, \bibinfo{person}{Qingsong Wen},
  \bibinfo{person}{Wei Ye}, \bibinfo{person}{Shikun Zhang}, {and}
  \bibinfo{person}{Yue Zhang}.} \bibinfo{year}{2024}\natexlab{}.
\newblock \bibinfo{title}{AutoSurvey: Large Language Models Can Automatically
  Write Surveys}.
\newblock
\showeprint[arxiv]{2406.10252}~[cs.IR]
\urldef\tempurl%
\url{https://arxiv.org/abs/2406.10252}
\showURL{%
\tempurl}


\bibitem[Weng et~al\mbox{.}(2025)]%
        {weng2024cycleresearcher}
\bibfield{author}{\bibinfo{person}{Yixuan Weng}, \bibinfo{person}{Minjun Zhu},
  \bibinfo{person}{Guangsheng Bao}, \bibinfo{person}{Hongbo Zhang},
  \bibinfo{person}{Jindong Wang}, \bibinfo{person}{Yue Zhang}, {and}
  \bibinfo{person}{Linyi Yang}.} \bibinfo{year}{2025}\natexlab{}.
\newblock \bibinfo{title}{CycleResearcher: Improving Automated Research via
  Automated Review}.
\newblock
\showeprint[arxiv]{2411.00816}~[cs.CL]
\urldef\tempurl%
\url{https://arxiv.org/abs/2411.00816}
\showURL{%
\tempurl}


\bibitem[Yang et~al\mbox{.}(2025)]%
        {yang2025qwen3technicalreport}
\bibfield{author}{\bibinfo{person}{An Yang}, \bibinfo{person}{Anfeng Li},
  \bibinfo{person}{Baosong Yang}, \bibinfo{person}{Beichen Zhang},
  \bibinfo{person}{Binyuan Hui}, \bibinfo{person}{Bo Zheng},
  \bibinfo{person}{Bowen Yu}, \bibinfo{person}{Chang Gao},
  \bibinfo{person}{Chengen Huang}, \bibinfo{person}{Chenxu Lv},
  \bibinfo{person}{Chujie Zheng}, \bibinfo{person}{Dayiheng Liu},
  \bibinfo{person}{Fan Zhou}, \bibinfo{person}{Fei Huang},
  \bibinfo{person}{Feng Hu}, \bibinfo{person}{Hao Ge}, \bibinfo{person}{Haoran
  Wei}, \bibinfo{person}{Huan Lin}, \bibinfo{person}{Jialong Tang},
  \bibinfo{person}{Jian Yang}, \bibinfo{person}{Jianhong Tu},
  \bibinfo{person}{Jianwei Zhang}, \bibinfo{person}{Jianxin Yang},
  \bibinfo{person}{Jiaxi Yang}, \bibinfo{person}{Jing Zhou},
  \bibinfo{person}{Jingren Zhou}, \bibinfo{person}{Junyang Lin},
  \bibinfo{person}{Kai Dang}, \bibinfo{person}{Keqin Bao},
  \bibinfo{person}{Kexin Yang}, \bibinfo{person}{Le Yu},
  \bibinfo{person}{Lianghao Deng}, \bibinfo{person}{Mei Li},
  \bibinfo{person}{Mingfeng Xue}, \bibinfo{person}{Mingze Li},
  \bibinfo{person}{Pei Zhang}, \bibinfo{person}{Peng Wang},
  \bibinfo{person}{Qin Zhu}, \bibinfo{person}{Rui Men}, \bibinfo{person}{Ruize
  Gao}, \bibinfo{person}{Shixuan Liu}, \bibinfo{person}{Shuang Luo},
  \bibinfo{person}{Tianhao Li}, \bibinfo{person}{Tianyi Tang},
  \bibinfo{person}{Wenbiao Yin}, \bibinfo{person}{Xingzhang Ren},
  \bibinfo{person}{Xinyu Wang}, \bibinfo{person}{Xinyu Zhang},
  \bibinfo{person}{Xuancheng Ren}, \bibinfo{person}{Yang Fan},
  \bibinfo{person}{Yang Su}, \bibinfo{person}{Yichang Zhang},
  \bibinfo{person}{Yinger Zhang}, \bibinfo{person}{Yu Wan},
  \bibinfo{person}{Yuqiong Liu}, \bibinfo{person}{Zekun Wang},
  \bibinfo{person}{Zeyu Cui}, \bibinfo{person}{Zhenru Zhang},
  \bibinfo{person}{Zhipeng Zhou}, {and} \bibinfo{person}{Zihan Qiu}.}
  \bibinfo{year}{2025}\natexlab{}.
\newblock \bibinfo{title}{Qwen3 Technical Report}.
\newblock
\showeprint[arxiv]{2505.09388}~[cs.CL]
\urldef\tempurl%
\url{https://arxiv.org/abs/2505.09388}
\showURL{%
\tempurl}


\bibitem[{Z.ai}(2025)]%
        {zai2025glm47}
\bibfield{author}{\bibinfo{person}{{Z.ai}}.} \bibinfo{year}{2025}\natexlab{}.
\newblock \bibinfo{title}{GLM-4.7: Advancing the Coding Capability}.
\newblock
\urldef\tempurl%
\url{https://z.ai/blog/glm-4.7}
\showURL{%
\tempurl}


\bibitem[Zhang et~al\mbox{.}(2019)]%
        {zhang2019oag}
\bibfield{author}{\bibinfo{person}{Fanjin Zhang}, \bibinfo{person}{Xiao Liu},
  \bibinfo{person}{Jie Tang}, \bibinfo{person}{Yuxiao Dong},
  \bibinfo{person}{Peiran Yao}, \bibinfo{person}{Jie Zhang},
  \bibinfo{person}{Xiaotao Gu}, \bibinfo{person}{Yan Wang},
  \bibinfo{person}{Bin Shao}, \bibinfo{person}{Rui Li}, {et~al\mbox{.}}}
  \bibinfo{year}{2019}\natexlab{}.
\newblock \showarticletitle{OAG: Toward linking large-scale heterogeneous
  entity graphs}. In \bibinfo{booktitle}{\emph{Proceedings of the 25th ACM
  SIGKDD international conference on knowledge discovery \& data mining}}.
  \bibinfo{pages}{2585--2595}.
\newblock


\bibitem[Zhang et~al\mbox{.}(2024)]%
        {zhang2024oag}
\bibfield{author}{\bibinfo{person}{Fanjin Zhang}, \bibinfo{person}{Shijie Shi},
  \bibinfo{person}{Yifan Zhu}, \bibinfo{person}{Bo Chen},
  \bibinfo{person}{Yukuo Cen}, \bibinfo{person}{Jifan Yu},
  \bibinfo{person}{Yelin Chen}, \bibinfo{person}{Lulu Wang},
  \bibinfo{person}{Qingfei Zhao}, \bibinfo{person}{Yuqing Cheng},
  {et~al\mbox{.}}} \bibinfo{year}{2024}\natexlab{}.
\newblock \showarticletitle{Oag-bench: a human-curated benchmark for academic
  graph mining}. In \bibinfo{booktitle}{\emph{Proceedings of the 30th ACM
  SIGKDD Conference on Knowledge Discovery and Data Mining}}.
  \bibinfo{pages}{6214--6225}.
\newblock


\bibitem[Zhang et~al\mbox{.}(2026)]%
        {zhang2026deepsurveybench}
\bibfield{author}{\bibinfo{person}{Guo-Biao Zhang}, \bibinfo{person}{Ding-Yuan
  Liu}, \bibinfo{person}{Da-Yi Wu}, \bibinfo{person}{Tian Lan},
  \bibinfo{person}{Heyan Huang}, \bibinfo{person}{Zhijing Wu}, {and}
  \bibinfo{person}{Xian-Ling Mao}.} \bibinfo{year}{2026}\natexlab{}.
\newblock \bibinfo{title}{DeepSurvey-Bench: Evaluating Academic Value of
  Automatically Generated Scientific Survey}.
\newblock
\showeprint[arxiv]{2601.15307}~[cs.AI]
\urldef\tempurl%
\url{https://arxiv.org/abs/2601.15307}
\showURL{%
\tempurl}


\bibitem[Zhang et~al\mbox{.}(2020)]%
        {zhang2020bertscoreevaluatingtextgeneration}
\bibfield{author}{\bibinfo{person}{Tianyi Zhang}, \bibinfo{person}{Varsha
  Kishore}, \bibinfo{person}{Felix Wu}, \bibinfo{person}{Kilian~Q. Weinberger},
  {and} \bibinfo{person}{Yoav Artzi}.} \bibinfo{year}{2020}\natexlab{}.
\newblock \bibinfo{title}{BERTScore: Evaluating Text Generation with BERT}.
\newblock
\showeprint[arxiv]{1904.09675}~[cs.CL]
\urldef\tempurl%
\url{https://arxiv.org/abs/1904.09675}
\showURL{%
\tempurl}


\bibitem[Zheng et~al\mbox{.}(2023)]%
        {zheng2023judgingllmasajudgemtbenchchatbot}
\bibfield{author}{\bibinfo{person}{Lianmin Zheng}, \bibinfo{person}{Wei-Lin
  Chiang}, \bibinfo{person}{Ying Sheng}, \bibinfo{person}{Siyuan Zhuang},
  \bibinfo{person}{Zhanghao Wu}, \bibinfo{person}{Yonghao Zhuang},
  \bibinfo{person}{Zi Lin}, \bibinfo{person}{Zhuohan Li},
  \bibinfo{person}{Dacheng Li}, \bibinfo{person}{Eric~P. Xing},
  \bibinfo{person}{Hao Zhang}, \bibinfo{person}{Joseph~E. Gonzalez}, {and}
  \bibinfo{person}{Ion Stoica}.} \bibinfo{year}{2023}\natexlab{}.
\newblock \bibinfo{title}{Judging LLM-as-a-Judge with MT-Bench and Chatbot
  Arena}.
\newblock
\showeprint[arxiv]{2306.05685}~[cs.CL]
\urldef\tempurl%
\url{https://arxiv.org/abs/2306.05685}
\showURL{%
\tempurl}


\bibitem[Zhu et~al\mbox{.}(2025)]%
        {zhu2025context}
\bibfield{author}{\bibinfo{person}{Kun Zhu}, \bibinfo{person}{Lizi Liao},
  \bibinfo{person}{Yuxuan Gu}, \bibinfo{person}{Lei Huang},
  \bibinfo{person}{Xiaocheng Feng}, {and} \bibinfo{person}{Bing Qin}.}
  \bibinfo{year}{2025}\natexlab{}.
\newblock \showarticletitle{Context-aware hierarchical taxonomy generation for
  scientific papers via llm-guided multi-aspect clustering}. In
  \bibinfo{booktitle}{\emph{Proceedings of the 2025 Conference on Empirical
  Methods in Natural Language Processing}}. \bibinfo{pages}{15627--15645}.
\newblock


\bibitem[Zhuang and Kennington(2024)]%
        {zhuang2024understandingsurveypapertaxonomy}
\bibfield{author}{\bibinfo{person}{Jun Zhuang} {and} \bibinfo{person}{Casey
  Kennington}.} \bibinfo{year}{2024}\natexlab{}.
\newblock \bibinfo{title}{Understanding Survey Paper Taxonomy about Large
  Language Models via Graph Representation Learning}.
\newblock
\showeprint[arxiv]{2402.10409}~[cs.CL]
\urldef\tempurl%
\url{https://arxiv.org/abs/2402.10409}
\showURL{%
\tempurl}


\end{thebibliography}
